\PassOptionsToPackage{table}{xcolor}
\documentclass{article}

\PassOptionsToPackage{numbers, compress}{natbib}
\usepackage[preprint]{neurips_2025}

\usepackage[utf8]{inputenc} % allow utf-8 input
\usepackage[T1]{fontenc}    % use 8-bit T1 fonts
\usepackage{hyperref}       % hyperlinks
\usepackage{url}            % simple URL typesetting
\usepackage{booktabs}       % professional-quality tables
\usepackage{amsfonts}       % blackboard math symbols
\usepackage{nicefrac}       % compact symbols for 1/2, etc.
\usepackage{microtype}      % microtypography
\usepackage{xcolor}         % colors
\usepackage{multirow}
\usepackage{listings}

\newcommand{\eg}{\textit{e.g.}}

\newcommand{\ie}{\textit{i.e.}}
\usepackage{pifont}% http://ctan.org/pkg/pifont
\newcommand{\cmark}{\ding{51}}%
\newcommand{\xmark}{\ding{55}}%

\usepackage{tikz}
\usetikzlibrary{positioning}
\usepackage{subcaption}
\usepackage{arydshln}
\usepackage{listings}
\usepackage{pgfplots}
\pgfplotsset{compat=1.18} 
\usepgfplotslibrary{groupplots}  % ✅ 正确加载 groupplots！
\usetikzlibrary{plotmarks}        % （如果需要标记形状）
\usetikzlibrary{shapes.geometric}
\usetikzlibrary{svg.path}
 
\title{Compile Scene Graphs with Reinforcement Learning}
\author{%
Zuyao Chen\textsuperscript{1, 2} \quad
Jinlin Wu\textsuperscript{3, 4} \quad
Zhen Lei\textsuperscript{3, 4} \quad
Marc Pollefeys\textsuperscript{2, 5} \quad
Chang Wen Chen\textsuperscript{1} \\
\textsuperscript{1}The Hong Kong Polytechnic University \quad
\textsuperscript{2}ETH Zürich \quad
\textsuperscript{3}CAIR, HKISI-CAS \\
\textsuperscript{4}Institute of Automation, CAS \quad
\textsuperscript{5}Microsoft \\
}
\usepackage{mathtools}

\begin{document}
\maketitle
%%%%%%%%%%%%%%%%%%%%%%%%%% 
\begin{abstract}
Next token prediction is the fundamental principle for training large language models (LLMs), 
and reinforcement learning (RL) further enhances their reasoning performance.
As an effective way to model language, image, video, and other modalities, 
the use of LLMs for end-to-end extraction of structured visual representations, such as scene graphs, remains underexplored.
It requires the model to accurately produce a set of objects and relationship triplets, rather than generating text token by token.
To achieve this, 
we introduce \emph{R1-SGG}, 
a multimodal LLM (M-LLM) initially trained via supervised fine-tuning (SFT) on the scene graph dataset and subsequently refined using reinforcement learning  to enhance its ability to generate scene graphs in an end-to-end manner.
The SFT follows a conventional prompt-response paradigm, 
while RL requires the design of effective reward signals. 
We design a set of graph centric rewards, including three recall based variants—Hard Recall, Hard Recall+Relax, and Soft Recall—which evaluate semantic and spatial alignment between predictions and ground truth at the object and relation levels.
A format consistency reward further ensures that outputs follow the expected structural schema.
Extensive experiments on the VG150 and PSG benchmarks show that R1-SGG substantially reduces failure rates and achieves strong performance in Recall and mean Recall, surpassing traditional SGG models and existing multimodal language models. 
%Our code is available at \url{https://github.com/gpt4vision/R1-SGG}.
\end{abstract}
%%%%%%%%%%%%%%%%%%%%%%
\section{Introduction}
Scene graphs, as structured visual representations, have gained increasing attention in many vision applications, 
such as robot manipulation~\citep{zhu2021hierarchical,zhang2025fungraph3d}, robot navigation~\citep{gu2024conceptgraphs, miao2024scenegraphloc,yin2024sg},  and medical image or video analysis~\citep{lin2022sgt,ozsoy20224d}, \textit{etc}. 
To generate scene graphs from an image, 
traditional Scene Graph Generation (SGG) models~\citep{johnson2015image,xu2017scene,DBLP:conf/iccv/LiOZWW17,zellers2018neural,tang2019learning,DBLP:conf/cvpr/ChenYCL19,li2022sgtr,DBLP:conf/nips/KhandelwalS22,DBLP:journals/pami/CongYR23, zhang2023learning, chen2024expanding} decouple the task into two subtasks, \ie, object detection and visual relationship recognition, 
and directly maximize the likelihood of the ground-truth labels given the image.
Essentially, these models tend to overfit the distribution of annotated datasets; 
Consequently, they struggle to handle long-tail distributions and are prone to generating biased scene graphs (\eg, all predicted relationships are head classes like ``on'' and ``of'').

While traditional SGG models rely on manual annotated datasets and struggle to generalize to new domains, 
recent advances in large languge models (LLMs) offer a new paragdim. 
LLM4SGG~\citep{kim2023llm4sgg} utilizes an LLM to extract relationship triplets from captions using both original and paraphrased text,  
while GPT4SGG~\citep{chen2023gpt4sgg} employs an LLM to synthesize scene graphs from dense region captions.  
Additionally, Li~\citep{li2024pixels} generates scene graphs via image-to-text generation using vision-language models (VLMs).  
These weakly supervised methods demonstrate potential for generating scene graphs with little or no human annotation  
but suffer from accuracy issues in the generated results.  
 
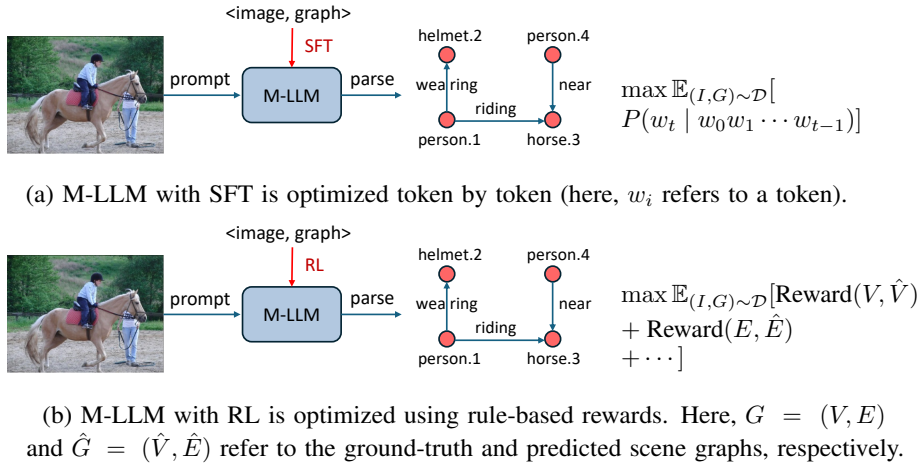
\begin{figure*}[htbp]
    \centering
    \resizebox{\textwidth}{!}{
    \begin{tikzpicture}
        % Traditional SGG
%        \node (a00) at(0, 0) {
%            \includegraphics[width=0.6\textwidth]{pics/figure-a}
%        };
%        \node (a01) [right=of a00, xshift=-10mm] {
%            $\max \mathbb{E}_{(I, G)\sim \mathcal{D}}{P(G \mid I)}$
%        };
%        \path (a00.west) -- (a01.east) coordinate[midway] (midcaption);
%        \node[below=of midcaption, align=center, yshift=0mm, text width=\textwidth] (captiontext-a) {
%            (a) Traditional SGG methods directly maximize the expectation of the likelihood  $ \mathbb{E}_{(I, G)\sim \mathcal{D}}{P(G \mid I)}$, where image-graph pairs $(I, G)$ are sampled from the dataset $\mathcal{D}$.
%        };    
        % SFT
        \node (a10) at(0, 0) {
            \includegraphics[width=0.6\textwidth]{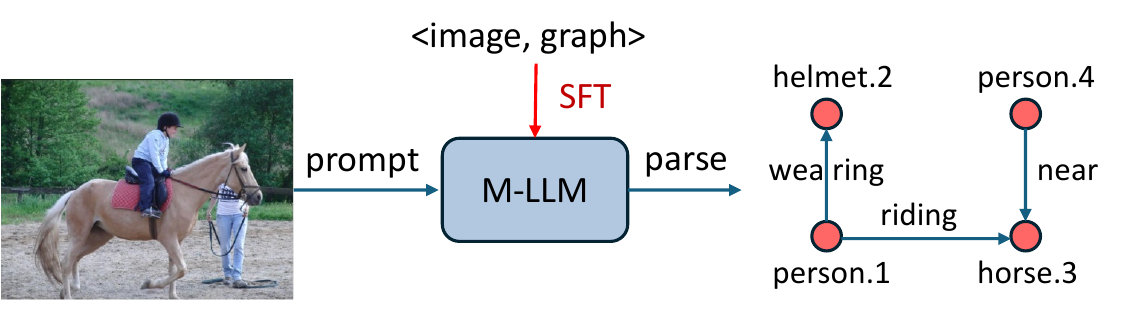}
        };
        \node (a11) [right=of a10, xshift=-10mm, yshift=-4mm, align=left] {
            $\max \mathbb{E}_{(I, G) \sim \mathcal{D}} [$\\
            $P(w_t \mid w_0 w_1 \cdots w_{t-1})]$
        };
        \path (a10.west) -- (a11.east) coordinate[midway] (midcaption2);
        \node[below=of midcaption2, align=center, yshift=-2mm, text width=\textwidth] (captiontext-b) {
            (a) M-LLM with SFT is optimized token by token (here, $w_i$ refers to a token).
        };    

        % RL
        \node (a20) [below=of a10, yshift=5mm] {
            \includegraphics[width=0.6\textwidth]{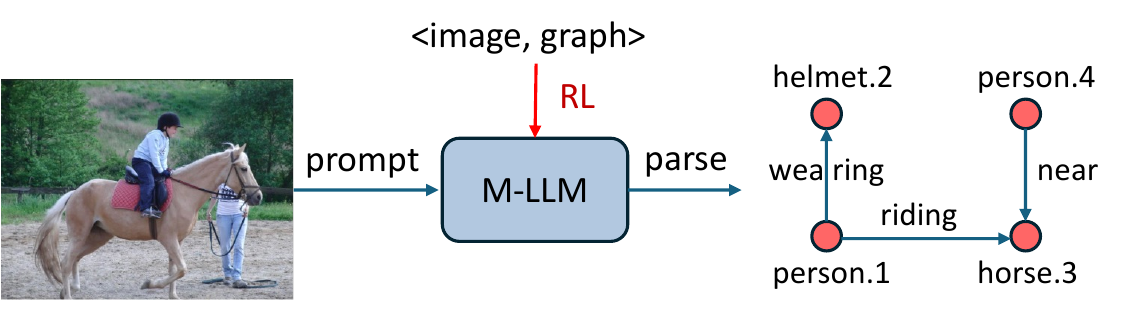}
        };
        \node (a21) [right=of a20, xshift=-10mm, yshift=-4mm, align=left] {
            $\max \mathbb{E}_{(I, G) \sim \mathcal{D}}[ \text{Reward}(V, \hat{V})$\\ 
            $+\ \text{Reward}(E, \hat{E})$ \\
            $+ \cdots ]$
        };       
        \path (a20.west) -- (a21.east) coordinate[midway] (midcaption3);
        \node[below=of midcaption3, align=center, yshift=-2mm, text width=\textwidth] (captiontext-c) {
            (b) M-LLM with RL is optimized using rule-based rewards. Here, $G = (V, E)$ and $\hat{G} = (\hat{V}, \hat{E})$ refer to the ground-truth and predicted scene graphs, respectively.
        };         
    \end{tikzpicture}
    }
\caption{Comparison of multimodal LLMs (M-LLMs) fine-tuned via Supervised Fine-tuning (SFT) and Reinforcement Learning (RL) for Scene Graph Generation (SGG).
}
    \label{fig:compare_para}
\end{figure*}
Despite these advancements, existing methods typically employ text-only LLMs or rely on intermediate captions as input, which do not fully leverage the rich visual context.
In contrast, multimodal large language models (M-LLMs) which integrate both visual and linguistic modalities offer the potential for more direct and holistic scene understanding. 
By processing visual information alongside natural language prompts, 
M-LLMs can generate scene graphs in an end-to-end manner.
However, in practice, 
M-LLMs suffer from instruction following (\eg, the output does not contain ``objects'' or ``relationships''), 
repeated response (\eg, 
\textit{\{"objects":[$\cdots$
\{"id": "desk.9", "bbox": [214, 326, 499, 389]\},
\{"id": "desk.10", "bbox": [214, 326, 499, 389]\},
\{"id": "desk.11", "bbox": [214, 326, 499, 389]\},
$\cdots$]}$\cdots$\} ),
inaccurate location, \textit{etc}.
These challenges highlight the need for better alignment between visual understanding and structured representation within the M-LLM framework.

To improve instruction-following and structured output generation in M-LLMs,
one intuitive solution is to perform Supervised Fine-tuning (SFT) on scene graph datasets (see Fig. \ref{fig:compare_para}-(a)). 
In the context of SGG, 
SFT aligns the model's outputs with expected formats (\eg, structured lists of objects and relationships) by training it on high-quality scene graph annotations.
This process encourages the model not only to recognize entities and relations from the image but also to organize them into a 
coherent and valid graph structure.
Nevertheless, SFT alone still be insufficient 
as all output tokens are weighted equally in the loss.
For example, the experimental results on the VG150 dataset~\cite{xu2017scene} reveal that even with SFT, M-LLM still has a high failure rate to generate a valid and high-quality scene graph. 
The drawback of SFT in SGG lies in the lack of effective signals to correct the output 
(\eg, the model cannot directly utilize the Intersection over Union (IoU)  between the predicted box and the ground truth to refine its output
).

To advance M-LLMs for effective Scene Graph Generation (SGG), 
we propose \emph{R1-SGG}, a novel framework leveraging visual instruction tuning enhanced by reinforcement learning (RL).
The visual instruction tuning stage follows a conventional supervised fine-tuning (SFT) paradigm, \ie, fine-tuning the model using prompt-response pairs with a cross-entropy loss. 
For the RL stage, we adopt GRPO, an online policy optimization algorithm introduced in DeepSeekMath~\cite{shao2024deepseekmath}.

To enable effective reinforcement learning for Scene Graph Generation,  we introduce a set of rule-based, graph-centric rewards that reflect the structural characteristics of scene graphs. 
Given an image and a prompt, a multimodal large language model (M-LLM) generates a set of objects and relational triplets. 
To evaluate and optimize these predictions, we formulate reward functions aligned with standard SGDET metrics~\citep{xu2017scene} and structured reasoning objectives. 
Specifically, we define three reward variants: \textbf{Hard Recall}, which counts a triplet as correct only if the subject, predicate, and object labels exactly match the ground truth and both bounding boxes achieve IoU > 0.5; 
\textbf{Hard Recall+Relax}, which relaxes the exact match constraint by incorporating embedding similarity between predicted and ground-truth labels; 
and \textbf{Soft Recall}, which further densifies reward signals via bipartite matching, combining object label similarity, IoU, and bounding box distance into a unified cost function.
These scene graph rewards are computed over matched object and edge pairs, and are complemented by a format reward that enforces structural adherence in output formatting. 
This reward design enables stable and fine-grained policy optimization using GRPO, guiding the M-LLM toward generating accurate, complete, and structurally valid scene graphs.

Our contributions can be summarized as follows: 
\begin{itemize}
    \item We explore how to develop a multimodal LLM for Scene Graph Generation (SGG), by leveraging visual instruction tuning with reinforcement learning (RL). To our knowledge, this is a pioneer work that develop a multimodal LLM to generate scene graphs in an end-to-end manner.
    \item  Graph-centric, rule-based rewards are designed to guide policy optimization in a manner aligned with standard evaluation metrics in SGG, such as the recall of relationship triplets—metrics that cannot be directly optimized through SFT.
    \item Experimental results demonstrate that the proposed framework improves the ability to understand and reason about scene graphs for multimodal LLMs. 
\end{itemize}
%%%%%%%%%%%%%%%%%%%%%%%% Related Work 
\section{Related Work}
\paragraph{Scene Graph Generation (SGG).}
Scene Graph Generation (SGG) is a foundational task in structured visual understanding, 
where the goal is to represent an image as a graph of objects and their pairwise relationships. 
Traditional approaches like \cite{xu2017scene,DBLP:conf/iccv/LiOZWW17,zellers2018neural,tang2019learning,DBLP:conf/cvpr/ChenYCL19,li2022sgtr,DBLP:conf/nips/KhandelwalS22,DBLP:journals/pami/CongYR23} decouple the task into object detection and relationship classification stages, 
and are typically trained via supervised learning on datasets such as Visual Genome (VG150)~\citep{xu2017scene}. 
While effective, these models are limited by their reliance on annotated data and exhibit strong bias toward head predicates such as ``on'' or ``of'', 
struggling on long-tail classes.

To overcome the closed-set limitation, recent work has explored open-vocabulary SGG.
For example, OvSGTR~\citep{chen2024expanding} extends scene graph prediction to a fully open-vocabulary setting by leveraging visual-concept alignment.
In parallel, weakly supervised methods have been developed to reduce the annotation burden.
These approaches, such as those proposed by \cite{zhong2021learning,li2022integrating,zhang2023learning,chen2024expanding}, use image-caption pairs as supervision to distill relational knowledge, enabling generalization to unseen concepts.

\paragraph{LLMs for Scene Graph Generation.}
With the rise of LLMs,
several studies have attempted to synthesize scene graphs from natural language. 
LLM4SGG~\citep{kim2023llm4sgg} extracts relational triplets from both original and paraphrased captions using text-only LLMs. 
GPT4SGG~\citep{chen2023gpt4sgg} goes a step further by using GPT-4 to generate scene graphs from dense region captions, improving contextual consistency and coverage. 
Meanwhile, \cite{li2024pixels} leverage vision-language models (VLMs) to produce scene graphs through image-to-text generation pipelines.

However, these caption-based or LLM-driven methods often exhibit limited accuracy, 
including incomplete object sets, and inconsistent relationship descriptions.
These issues arise from the lack of structure in the generated outputs and the absence of mechanisms to refine the results according to scene-level constraints.

\paragraph{Reinforcement Learning (RL) for LLMs.}
Reinforcement learning (RL) has been increasingly adopted to enhance the reasoning capabilities of large models. 
Algorithms like Proximal Policy Optimization (PPO)~\citep{schulman2017proximal}
and Group Relative Policy Optimization (GRPO)~\citep{shao2024deepseekmath} guide models using reward signals instead of relying solely on maximum likelihood estimation. 
In the context of large language models, DeepSeek-R1~\citep{guo2025deepseek} demonstrates that RL can significantly improve structured reasoning and planning.

In multimodal learning, however, RL remains underutilized for generating structured outputs.
Our work addresses this by introducing rule-based reward functions at multiple levels, including three scene graph reward variants and a format consistency reward.
These signals promote the generation of meaningful and coherent scene graphs by explicitly evaluating alignment with ground-truth annotations.
%%%%%%%%%%%%%%%%%%%%%%%%%%%%% 
\section{Methodology}
\subsection{Preliminary}
\paragraph{Scene Graph Generation (SGG).}
Scene graph generation (SGG) transforms an image $I$ into a structured representation that captures both objects and their interactions. Specifically, SGG produces a directed graph $\mathcal{G}=(\mathcal{V}, \mathcal{E})$, where each node $v_i \in \mathcal{V}$ represents an object annotated with an object category $c_i$ and a bounding box $b_i$. Each relationship triplet $e_{ij} \in \mathcal{E}$ captures the relationship between two nodes. The triplet is defined as 
$
e_{ij} := \langle v_i,\, p_{ij},\, v_j \rangle,
$
where $p_{ij}$ encodes the visual relationship between the subject $v_i$ and the object $v_j$, such as spatial relations (\eg, ``on'', ``under'') or interactive relations (\eg, ``riding'', ``holding''). Typically, SGG models decouple this task into two subtasks, namely object detection and relationship recognition, both optimized by maximizing the likelihood of the corresponding ground-truth labels given the image.

\paragraph{Reinforcement Learning with GRPO.}
Group Relative Policy Optimization (GRPO) is a online reinforcement learning algorithm introduced by DeepSeekMath~\citep{shao2024deepseekmath}. Unlike traditional methods such as PPO~\citep{schulman2017proximal}, which require an explicit critic network, GRPO instead compares groups of candiates to update the policy $\pi_{\theta}$.
Specifically, for each input query $q$, a set of candidate outputs \(\{o_i\}_{i=1}^{G}\) is drawn from the previous policy \(\pi^{\text{old}}(O|q)\), and the advantage of each candidate is computed relative to the group's average reward:
\begin{equation}
A_i = \frac{r_i - \mathrm{mean}(\{r_1, \dots, r_G\})}{\mathrm{std}(\{r_1, \dots, r_G\})}.
\end{equation}
The policy parameters $\theta$ are updated by maximizing the following GRPO objective:
\begin{equation}
\begin{aligned}
J_{\text{GRPO}}(\theta) = \; & \mathbb{E}_{q \sim P(Q),\, \{o_i\}_{i=1}^{G} \sim \pi^{\text{old}}(O|q)}\Bigg[\frac{1}{G}\sum_{i=1}^{G}\Bigg(
\min\Bigg(\frac{\pi_\theta(o_i|q)}{\pi^{\text{old}}(o_i|q)}A_i,\,\\
&\text{clip}\Big(\frac{\pi_\theta(o_i|q)}{\pi^{\text{old}}(o_i|q)},\,1-\epsilon,\,1+\epsilon\Big)A_i\Bigg)
- \beta\, D_{\text{KL}}\big(\pi_\theta\parallel\pi_{\text{ref}}\big)\Bigg],
\end{aligned}
\label{eq:grpo}
\end{equation}
Here, $\epsilon$ and $\beta$ are hyper-parameters.
The first term uses a clipped probability ratio (as in PPO) to control the update magnitude, 
while the KL divergence regularizer $D_{\text{KL}}(\pi_\theta \parallel \pi_{\text{ref}})$ constrains the new policy $\pi_\theta$ to not deviate too much from a reference policy $\pi_{\text{ref}}$.
This formulation, 
which combines a group-relative advantage, a clipping mechanism, and a KL divergence regularizer, 
stabilizes policy updates and improves training efficiency, 
demonstrating remarkable potential for enhancing the reasoning performance of LLMs such as DeepSeek R1~\citep{guo2025deepseek}.
%%%%%%%%%%%%%%%%%%%%%%%%%%%%%%%%%%%
\subsection{Overview of R1-SGG}
R1-SGG is a reinforcement learning framework that enhances scene graph generation (SGG) in multimodal large language models (M-LLMs). 
It builds on a supervised fine-tuning (SFT) stage using prompt-response pairs, followed by reinforcement learning (RL) with structured, graph-centric rewards.

Given an input image and prompt, the M-LLM generates a scene graph \(\mathcal{G}_{\text{pred}} = (\mathcal{V}_{\text{pred}}, \mathcal{E}_{\text{pred}})\), comprising objects (nodes) and their relationships (edges). 
We primarily optimize using \emph{Hard Recall}, which aligns with SGDET metrics by rewarding exact triplet matches. 
To study the sparsity and design of the rewards, we also evaluated relaxed alternatives based on bipartite matching between \(\mathcal{G}_{\text{pred}}\) and the ground truth graph \(\mathcal{G}_{\text{gt}}\), allowing fine-grained node and edge rewards.
Our RL pipeline employs Group Relative Policy Optimization (GRPO)~\citep{shao2024deepseekmath}, which compares sampled outputs and promotes higher-reward candidates. 
By integrating SFT, GRPO, and graph-aware rewards, R1-SGG enables M-LLMs to generate accurate, diverse, and structurally valid scene graphs.

\subsection{Rewards Definition}
\label{sec:rewards}
\subsubsection{Format Reward}
Following DeepSeek R1~\citep{guo2025deepseek}, 
we employ a format reward to ensure that the model's response adheres to the expected structure, 
specifically \texttt{<think>$\cdots$</think><answer>$\cdots$</answer>}. 
A reward of 1 is assigned if the response follows this format 
and the segment enclosed by \texttt{<answer>$\cdots$</answer>} contains both the keywords \texttt{"object"} and \texttt{"relationships"}; otherwise, the reward is 0.
%%%%%%%
\subsubsection{Scene Graph Rewards}
Standard evaluation protocols for Scene Graph Generation (SGG), such as SGDET~\citep{xu2017scene}, formulate the task as a recall-oriented problem, 
emphasizing the model's ability to retrieve correct relationship triplets from an image.
To investigate the impact of different reward formulations, we introduce three variants: \emph{Hard Recall}, \emph{Hard Recall+Relax}, and \emph{Soft Recall}.

\paragraph{Hard Recall.}
To align policy optimization with standard SGDET metrics, we define \emph{Hard Recall}, where a predicted triplet 
\(\langle \texttt{subject}, \texttt{predicate}, \texttt{object} \rangle\) 
is counted as a true positive when \emph{both} of the following hold:
1) \emph{Triplet accuracy:} the subject, predicate, and object labels exactly match the ground truth.
2) \emph{Localization accuracy:} the IoU between predicted and ground-truth bounding boxes exceeds 0.5.

This reward is aligned with standard metrics but suffers from sparsity due to its strict criteria.

\paragraph{Hard Recall + Relax.}
We relax the triplet accuracy requirement by computing cosine similarity between the entity embeddings of predicted and ground-truth triplets. 
This softens the discrete matching constraint to provide more gradient signal.

\paragraph{Soft Recall.}
We further propose a dense matching reward by formulating it as a bipartite matching problem, similar to DETR~\citep{carion2020end}, where predicted nodes \(\{v_i = (c_i, b_i) \}_{i=1}^{M}\) (each node $v_i$ is comprising an object class $c_i$ and a bounding box $b_i$) are matched to ground-truth nodes \(\{\tilde{v}_j = (\tilde{c}_j, \tilde{b}_j) \}_{j=1}^{N}\) with the following cost:
\begin{equation}
\begin{aligned}
    \text{cost}(v_i, \tilde{v}_j) = &\lambda_1 \cdot (1.0 - \langle \texttt{Embedding}(c_i),
    \texttt{Embedding}(\tilde{c}_j) \rangle) \\
    & + \lambda_2 \cdot 
    (1.0 - \mathrm{IoU}(b_i, \tilde{b}_j)) + 
    \lambda_3 \cdot ||b_i - \tilde{b}_j||_1,
\end{aligned}
\end{equation}
where $\langle \cdot, \cdot \rangle$ denotes cosine similarity, 
$\lambda_1$, $\lambda_2$ are weight factors, 
and \texttt{Embedding} is obtained via the NLP tool SpaCy.
By solving the bipartite matching problem, 
we establish a one-to-one node matching between the predicted graph $\mathcal{G}_{\text{pred}}$ and the ground-truth graph $\mathcal{G}$.

For a predicted node $v_i$,
the reward is defined as 
\begin{equation}
    \text{Reward}(v_i) = 
    \begin{cases} 
        \lambda_1 \cdot \left\langle \texttt{Embedding}(c_i), \texttt{Embedding}(\tilde{c}_j) \right\rangle  
        \\
        \quad + \lambda_2 \cdot \text{IoU}(b_i, \tilde{b}_j) \\
        \quad + \lambda_3 \cdot \exp(-||b_i - \tilde{b}_j||_1), 
        & \text{if } v_i \text{ and } \tilde{v}_j \text{ are matched}, \\
        0, & \text{otherwise}.
    \end{cases}
\end{equation}
which is the linear combination of object category similarity and the IoU of bounding boxes.
The total rewards of an image's prediction set 
$\{v_i\}_{i=1}^{M}$
is computed as 
\begin{equation}
 \text{Reward}(\{v_i\}_{i=1}^{M}) = \frac{1}{|\mathcal{V}_{\mathrm{gt}}|} \sum_{i=1}^{M} \text{Reward}(v_i) . 
\end{equation}
 
For a predicted triplet $e_{ij} := <v_i, p_{ij}, v_j>$, 
the reward is defined as 
\begin{equation}
\text{Reward}(e_{ij})=
\begin{cases} 
\langle \texttt{Embedding}(v_{i}),\,
\texttt{Embedding}(\tilde{v}_{k}) \rangle \cdot \\ 
\langle \texttt{Embedding}(v_{j}),\, 
\texttt{Embedding}(\tilde{v}_{l}) \rangle \cdot \\
\langle \texttt{Embedding}(p_{ij}),\,
\texttt{Embedding}(p_{kl}) \rangle, & \text{if } v_i \text{ matches } \tilde{v}_k \\
& \quad \text{and } v_j \text{ matches } \tilde{v}_l, \\
0, & \text{otherwise.}
\end{cases}
\end{equation}
Thereby, the reward of an image's predicted edge set 
is computed as 
\begin{equation}
    \text{Reward}(\{e_{ij} \}) = 
    \frac{1}{|\mathcal{E}_{\mathrm{gt}} |} \sum \text{Reward}(e_{ij}) .
\end{equation}

\section{Experiments} 
\subsection{Dataset and Experiment Setup}
\label{sec:exp_setting}
\textbf{Dataset.} 
The widely-used scene graph dataset VG150~\cite{xu2017scene} consists of 150 object categories and 50 relation categories.
Following prior works~\citep{zhang2023learning,chen2024expanding}, the training set used in this work contains 56,224 image-graph pairs, while the validation set includes 5,000 pairs.
To prompt the M-LLM, we transform each image-graph pair using the template described in Table~\ref{tab:prompt_sg}.

The Panoptic Scene Graph (PSG) dataset~\cite{yang2022panoptic} is built on the COCO dataset~\cite{lin2014microsoft},  
consisting of 80 \emph{thing} object categories, 53 \emph{stuff} object categories, and 56 relation categories.
It contains 46,563 image-graph pairs for training and 2,186 pairs for testing.

\textbf{Evaluation.}
Following the standard evaluation pipeline in SGG, 
we adopt the SGDET protocol~\citep{xu2017scene,tang2020unbiased} to measure the model's ability to generate scene graphs.
SGDET requires the model to generate scene graphs directly from the image without any predefined object boxes. 
Performance is evaluated using Recall and mean Recall (mRecall).
Recall is computed for each image-graph pair, where a predicted triplet is considered correct if both the subject and object bounding boxes have an Intersection over Union (IoU) of at least 0.5 with the corresponding ground-truth boxes, and the subject category, object category, and relationship label all match the ground truth.
Mean Recall (mRecall) is obtained by averaging the Recall across all relation categories.
We additionally report AP@50 to assess object detection performance and Failure Rate to evaluate format consistency.

\textbf{Implementation Details.}
Our code is based on the \texttt{trl} library~\cite{vonwerra2022trl} and utilizes vLLM~\cite{kwon2023efficient} to speed up sampling during reinforcement learning.
For SFT, the model is trained for 3 epochs with a batch size of 128 on 4 NVIDIA A100 (80GB) GPUs, 
using the AdamW optimizer~\cite{adamw} with a maximum learning rate of 1e-5. 
For RL, the model is trained for 1 epoch with a batch size of 32 and 8 generations per sample on 16 NVIDIA GH200 (120GB) GPUs, 
also using AdamW with a maximum learning rate of 6e-7.
%SFT training takes approximately 4 hours for 2B models and 10 hours for 7B models, while RL training requires about 12 hours (2B) and 16 hours (7B).

\subsection{How Well Do M-LLMs Reason About Visual Relationships?}
We evaluate the visual relationship reasoning capabilities of open-source multimodal LLMs using a four-to-one Visual Question Answering (VQA) task. 
Each model is prompted with an image and a corresponding question. 
The used prompt template is: 
\texttt{
Analyze the relationship between the object "\{sub\_name\}" at \{sub\_box\} and the object "\{obj\_name\}" at \{obj\_box\} in an image of size (\{width\}x\{height\}). 
The bounding boxes are in [x1, y1, x2, y2] format. 
Choose the most appropriate relationship from the following options: 
A) \{choices[0]\}; 
B) \{choices[1]\}; 
C) \{choices[2]\}; 
D) \{choices[3]\}.
}
We report Acc (accuracy over all questions) and mAcc (mean accuracy per image) in Table~\ref{tab:accuracy_comparison}.
The results reveal that many multimodal LLMs struggle with visual relationship reasoning. 
Moreover, the task exhibits a noticeable text bias, and the presence of bounding boxes can sometimes mislead the model's attention.
As a simpler task compared to SGG, the poor performance suggests that directly applying multimodal LLMs to SGG may yield suboptimal results.

\begin{table}[t]
    \centering
\caption{SGDET performance on the VG150 validation set. For M-LLMs, predefined object classes and relation categories are included in the input prompts.}
    \resizebox{0.8\textwidth}{!}{
    \begin{tabular}{lccccc}
    \toprule
    \textbf{Method} & \textbf{Params}  & \textbf{Failure Rate (\%)} & \textbf{AP@50} & \textbf{Recall} & \textbf{mRecall} \\
    \midrule
    \multicolumn{6}{l}{\emph{Specific Models}} \\
    IMP~\citep{xu2017scene} &  \multirow{4}{*}{-} & \multirow{4}{*}{-}  & 20.91 & 17.85 &  2.66\\ 
    MOTIFS~\citep{zellers2018neural} &  & &29.56   & 27.21   &  7.84 \\
    VCTree~\citep{tang2019learning} &  & & 28.13 &24.87  & 8.47  \\
    OvSGTR~\citep{chen2024expanding} &  &  & 33.39 & 26.74 & 5.83 \\ 
    \midrule 
    \multicolumn{6}{l}{\emph{Commercial M-LLMs}} \\
    GPT-4o~\citep{hurst2024gpt} & - & 2.94 & 0.00 & 0.00 & 0.00  \\ 
    Gemini 1.5 Flash~\citep{reid2024gemini} & - & 1.10 & 0.51 & 0.10 & 0.08\\
    Gemini 2.0 Flash~\citep{google_gemini2_flash} & - &  1.06 & 0.54 & 0.07 & 0.03\\   
    \midrule 
    \multicolumn{6}{l}{\emph{Open-sourced M-LLMs}} \\
    LLaVA v1.5~\citep{liu2023improvedllava} & 7B &  82.70 & 0.00 & 0.00 & 0.00 \\
 %   ASMv2~\citep{wang2024all} & 13B &   \\
 Qwen2-VL-2B-Instruct~\citep{Qwen2VL}   & 2B &  59.96 & 2.18  & 0.07 & 0.18 \\
     \quad+SFT        & 2B &  72.42 & 8.10 & 5.47 & 1.46    \\
     Qwen2-VL-7B-Instruct~\citep{Qwen2VL}    & 7B &  54.46 &   6.07 & 0.69 & 0.80 \\
    \quad+SFT        & 7B & 39.54 & 14.18 & 9.62 & 3.30\\
    \hdashline    
\rowcolor{lightgray!30}     R1-SGG-Zero    & 2B & 0.34 & 12.30 & 11.89 & 5.70        \\   
\rowcolor{lightgray!30}      R1-SGG     & 2B &    0.10  & 17.87 & 21.09 & 7.48  \\
\rowcolor{lightgray!30}      R1-SGG-Zero    & 7B &  0.04 & 15.59 & 18.34 & 8.32    \\    
 \rowcolor{lightgray!30}     R1-SGG     & 7B &   \textbf{0.08}   & \textbf{19.47}   & \textbf{23.75}  & 
    \textbf{11.43}  \\ 
    \bottomrule 
    \end{tabular}
}
    \label{tab:qwen2vl}
\end{table}
\subsection{How Well do M-LLMs Generate Scene Graphs?}
\subsubsection{Benchmark on VG150}
We report the performance under various settings in Table~\ref{tab:qwen2vl}, which includes:
1) \emph{Specific Models}: Methods built on specific detectors such as Faster R-CNN~\citep{ren2015faster} (e.g., IMP~\citep{xu2017scene}) or DETR~\citep{carion2020end} (e.g., OvSGTR~\citep{chen2024expanding}) for scene graph generation.
2) \emph{Commercial M-LLMs}: Advanced multimodal large language models such as GPT-4o~\citep{hurst2024gpt} and Gemini~1.5~Flash~\citep{reid2024gemini}.
3) \emph{Open-source M-LLMs}: Publicly available models such as LLaVA~v1.5~\citep{liu2023improvedllava}, Qwen2-VL~\citep{Qwen2VL}, and our proposed \emph{R1-SGG-Zero} (based on \texttt{Qwen2-VL-2B/7B-Instruct}, trained with GRPO but without supervised fine-tuning) and \emph{R1-SGG} (built on the same backbone, fine-tuned with GRPO and initialized from SFT checkpoints).

The results in Table~\ref{tab:qwen2vl} reveal several key observations.

\textbf{Zero-shot Performance of M-LLMs.}
Either commercial or open-source multimodal LLMs struggle to generate accurate scene graphs, and this can be attributed to several factors. 
First, the internal processing of private models such as GPT-4o remains opaque to users, resulting in suboptimal object detection performance. 
Second, models like LLaVA~v1.5 align visual and textual features only at the image level, typically using a fixed resolution of 336$\times$336, which restricts spatial understanding. 
Third, although models such as Gemini~2.0 and Qwen2-VL demonstrate a degree of spatial understanding, 
the task of scene graph generation is much complex than pure object detection or visual grounding. 
Consequently, their zero-shot performance drops significantly.

\textbf{SFT vs. RL.}
\textbf{1)} RL substantially improves performance across all metrics compared to SFT alone. Specifically, RL dramatically reduces the failure rate (\eg, from 72.42\% to 0.10\% for 2B models) and yields significant gains in AP@50, Recall, and mRecall. This highlights the effectiveness of GRPO in enhancing the model's ability to generate accurate and complete scene graphs.
\textbf{2)} SFT achieves moderate improvements in AP@50 and Recall over the baseline but struggles with a relatively high failure rate. This suggests that SFT primarily improves relation prediction while being less effective at correcting structural errors, such as missing objects, relationships, or format inconsistencies.
\textbf{3)} applying RL on top of SFT (\ie, R1-SGG) further boosts performance over both SFT and R1-SGG-Zero in most cases. This indicates that combining SFT and RL benefits from better initialization, 
leading to stronger relation recognition and higher recall.
\textbf{4)} larger models (\eg, 7B) consistently outperform smaller models (\eg, 2B) across AP@50, Recall, and mRecall, demonstrating the benefits of scaling model capacity for scene graph generation.

\textbf{Compared to Specific Models.}
The gap between AP@50 and Recall highlights the advantage of dense predictions. 
However, our models, such as \emph{R1-SGG}, achieve a notable mean Recall (mRecall) of 11.43\%, 
suggesting that multimodal LLMs are more effective at generating less biased scene graphs.
Moreover, specific models are typically restricted to a limited vocabulary and struggle to generalize across domains, 
whereas multimodal LLMs exhibit greater adaptability and broader generalization capabilities.

Overall, the results demonstrate that reinforcement learning (RL) significantly reduces the failure rate and enhances both object detection and relationship recognition. In contrast, supervised fine-tuning (SFT) alone results in a relatively high failure rate and limited improvements. As shown in Fig.~\ref{fig:zero_rl_metrics_steps}, the failure rate quickly drops to near-zero with RL, whereas SFT continues to suffer from frequent structural errors. 
%Meanwhile, our experimental results suggest that predefined object and relation categories are unnecessary, despite their potential to reduce the search space of M-LLMs.
\subsubsection{Benchmark on PSG}
\begin{table}[t]
\centering
\caption{
Performance on the PSG dataset~\citep{yang2022panoptic}. 
For M-LLMs, predefined object classes and relation categories are included in the input prompts.
%Note that $\text{ASMv2}^*$~\citep{wang2024all} originally reports results on 1,000 samples, 
%while all other methods use the full test set of 2,186 samples. 
%Here, we correct this discrepancy for fair comparison.
}
\resizebox{0.8\textwidth}{!}{
\begin{tabular}{lccccc}
\toprule
\textbf{Model} & \textbf{Params} & \textbf{Failure Rate  (\%)} & \textbf{AP@50} & \textbf{Recall} & \textbf{mRecall} \\
\midrule
\multicolumn{6}{l}{\emph{Specific Models}} \\
IMP~\citep{xu2017scene} &  &  &   & 16.50 & 6.50 \\
MOTIFS~\citep{zellers2018neural} &  &  &   & 20.00 & 9.10 \\
VCTree~\citep{tang2019learning} & - & - & -  & 20.60 & 9.70 \\
GPSNet~\citep{lin2020gps} &   &  &   & 17.80 & 7.00 \\
PSGFormer~\citep{yang2022panoptic} &  &  &   & 18.60 & 16.70 \\
\midrule
%\multicolumn{6}{l}{\emph{Commercial M-LLMs}} \\
%Gemini 1.5 Flash & \\
%GPT-4o & \\ 
%\midrule
\multicolumn{6}{l}{\emph{Open-sourced M-LLMs}} \\
LLaVA v1.5~\citep{liu2023improvedllava} & 7B & 81.97 & 0.07 & 0.00 & 0.00 \\
TextPSG~\citep{zhao2023textpsg} & - & -  & -  & 4.80 & - \\
%TextPSG~\citep{zhao2023textpsg} & - & -  & -  & 5.50 & - \\
%$\text{ASMv2}^*$~\citep{wang2024all} & 13B &  - &  - &  14.20 & 10.30 \\
ASMv2~\citep{wang2024all} & 13B &  0.87  & 21.45  &   14.77 & 11.82  \\
LLaVA-SpaceSGG~\citep{xu2025llavaspacesgg} & 13B &  - & - & 15.43 & 13.23 \\ 

 Qwen2-VL-2B-Instruct   & 2B &  67.20 &  4.89 & 0.39 & 0.26 \\
     \quad+SFT        & 2B &  6.54 & 36.05 & 22.06 & 14.92 \\ 
Qwen2-VL-7B-Instruct    & 7B & 37.97 & 12.75 & 3.18 & 4.33   \\
    \quad+SFT        & 7B & 0.96 & 40.79 & 24.73 & 17.11 \\
 \hdashline 
\rowcolor{lightgray!30}     R1-SGG-Zero    & 2B &  
 0.23 & 25.61 & 25.06 & 18.15
\\   
\rowcolor{lightgray!30}      R1-SGG     & 2B &   
2.70 & 39.28 & 38.49 & 31.21 
\\
\rowcolor{lightgray!30}      R1-SGG-Zero    & 7B &  0.00 & 32.92 & 37.00 & 32.04   \\    
\rowcolor{lightgray!30}      R1-SGG     & 7B &  \textbf{0.00} & \textbf{42.05} & \textbf{43.48} & \textbf{33.71}   \\ 
\bottomrule
\end{tabular}
}
\label{tab:psg}
\end{table}
 As shown in Table~\ref{tab:psg}, our R1-SGG approach achieves strong performance on the PSG dataset. 
 Compared to baselines, SFT significantly improves AP@50, Recall, and mean Recall (mRecall), while reinforcement learning further enhances relationship recognition, achieving the highest Recall (43.48\% for 7B model) and mRecall (33.71\%). 
 Notably, our method also drives the failure rate to zero, demonstrating the effectiveness of reinforcement learning in promoting structured, accurate scene graph generation even without predefined object categories.

\subsection{Qualitative Results}
We present qualitative results in Fig.~\ref{fig:qual_vis_vg} and Fig.~\ref{fig:qual_vis_psg}. 
As shown in Fig.~\ref{fig:qual_vis_vg}, the ground-truth scene graph (Fig.~\ref{fig:qual_vis_vg}-(a)) captures key objects and their relationships but is biased toward the predicate ``\emph{has}''. 
Conversely, the zero-shot {Qwen2-VL-7B-Instruct} (Fig.~\ref{fig:qual_vis_vg}-(b)) fails to generate a valid JSON output, indicating poor instruction-following ability. 
With supervised fine-tuning, the model produces structurally valid graphs (Fig.~\ref{fig:qual_vis_vg}-(c)) but frequently omits important relationships, resulting in a sparse scene graph. 
R1-SGG-Zero (7B), trained with RL only, improves relational recall and structure (Fig.~\ref{fig:qual_vis_vg}-(d)), yet still outputs inaccurate triplets such as \emph{$<$wheel, on, horse$>$} and \emph{$<$helmet.2, on, horse$>$}. 
Finally, R1-SGG (7B), trained with both SFT and RL, produces a complete and consistent scene graph (Fig.~\ref{fig:qual_vis_vg}-(e)), with results that even surpass the ground truth in relational richness.

\subsection{Discussion}
\label{sec:discussion}
Through the exploration of applying GRPO to the SGG task, we make several observations.

\noindent\textbf{KL Regularization.} We compare models trained with and without KL divergence regularization in Fig.~\ref{fig:kl_sample_group_vs_steps}. 
From the result, removing KL regularization leads to improved performance, particularly with a significant reduction in failure rate.

\noindent\textbf{Sampling Length.} 
In our experiments, the default sampling length is set to 1,024, which sufficiently covers most corrected answers. 
As shown in Fig.~\ref{fig:kl_sample_group_vs_steps}, increasing the sampling length to 2,048 does not yield further performance improvements, 
suggesting that longer sampling might enlarge the search space and introduce additional optimization difficulties without clear benefits.
This observation aligns with prior findings on test-time scaling, where increasing Chain-of-Thought (CoT) length can degrade performance~\citep{zeng2025revisiting}.

\noindent\textbf{Group Size.}
As shown in Fig.~\ref{fig:kl_sample_group_vs_steps}, increasing the group size from 8 to 16 stabilizes training performance, consistent with the intuition that more candidates reduce variance in group statistics estimation. 
To balance computational cost and performance, we adopt a group size of 8 as the default in this work.

\noindent\textbf{To Think or Not Think?}
We adopt the \texttt{<think>$\cdots$</think><answer>$\cdots$</answer>} format in the system prompt, following DeepSeek R1~\citep{guo2025deepseek}.
However, models such as \texttt{Qwen2-VL-2B/7B-Instruct} often fail to produce outputs with the \texttt{<think>} tag after fine-tuning, indicating difficulty in adhering to the intended structure. 
This suggests that rule-based rewards alone are insufficient to trigger abstract reasoning patterns like CoT, and highlights the need for additional SFT on CoT-specific datasets to incentivize coherent intermediate reasoning.

\noindent\textbf{Generalization Across Datasets.}
\begin{table}[t]
    \centering
\caption{Generalization across datasets using \texttt{Qwen2-VL-7B-Instruct} as the baseline. 
Columns under \emph{Pre-training} indicate whether the weights were initialized from specific checkpoints, 
while the \emph{Training} column specifies the dataset(s) used during the fine-tuning stage.
``\emph{w/o} cats.'' denotes prompts without predefined object classes or relation categories.
}
    \resizebox{\textwidth}{!}{
    \begin{tabular}{lcccccccccc}
        \toprule
        \multirow{2}{*}{\textbf{Model}} & \multirow{2}{*}{\textbf{Pre-Training}}& \multirow{2}{*}{\textbf{Training}} & \multicolumn{4}{c}{\textbf{VG150}} & \multicolumn{4}{c}{\textbf{PSG}} \\
        \cmidrule(r){4-7} \cmidrule(l){8-11}
        & & &\textbf{Failure Rate} & \textbf{AP@50} & \textbf{Recall} & \textbf{mRecall} & \textbf{Failure Rate} & \textbf{AP@50} & \textbf{Recall} & \textbf{mRecall} \\
        \midrule 
        % Example row
        baseline  & - & - & 
        54.46 & 6.07 & 0.69  & 0.80
        & 37.97 & 12.75 & 3.18 & 4.33
        \\    
        baseline (\emph{w/o} cats.) & - & - & 44.58 & 6.83 &  0.61 & 0.37 & 30.28 & 13.79 & 1.96 & 2.30 \\
        SFT & - & VG150 & 39.54 & 14.18 & 9.62 & 3.30 & 
        22.10 & 11.05 & 3.03 & 1.36\\ 
        SFT (\emph{w/o} cats.) & - & VG150 &  42.98 & 13.03 & 8.94 & 2.47 &    19.81 & 12.15 & 3.87 & 1.81 \\
        R1-SGG-Zero & - & VG150 &  
        0.04 & 15.59 & 18.34 & 8.32 
        & \textbf{0.18} & \textbf{24.92} & \textbf{13.83}  & \textbf{8.90} 
        \\
        R1-SGG-Zero (\emph{w/o} cats.) & - &VG150 &  
        0.06 & 15.30 & 16.33 & 6.94 &
        0.18 & 18.10 & 6.16 & 3.38
        \\  
         R1-SGG  & SFT   & VG150 & 
         \textbf{0.08}   &  \textbf{19.47}   &  \textbf{23.75}  & \textbf{11.43} 
         & {0.23} & {18.12} & {9.10} & {5.13} 
         \\       
        R1-SGG (\emph{w/o} cats.) & SFT (\emph{w/o} cats.)&VG150 &  0.30 & 18.09 & 22.73 & 9.62 &
        0.64 & 14.64 & 7.51 & 3.88 
        \\    
        \hdashline 
         SFT  & - & PSG &  
         36.98 & 5.79 & 1.42 & 0.77 & 
         0.91 & 40.58 & 24.75 & 17.31 \\          
        SFT (\emph{w/o} cats.)& - & PSG &    2.54&  7.94 &  1.77 & 1.25  &   1.01 & 39.02 & 23.70 & 17.17   \\   
        R1-SGG-Zero  & - & PSG &
        \textbf{0.12} &  \textbf{14.22} &  \textbf{8.90} & \textbf{5.34} & 
        0.00 & 32.92 & 37.00 & 32.04\\     
        R1-SGG-Zero (\emph{w/o} cats.) & - &PSG &   
        0.02 & 9.08 & 2.80 & 1.78 &
        0.05 & 24.26 & 19.94 & 18.04
         \\
        R1-SGG  & SFT   & PSG&
        0.94& 10.38 & 4.40 & 2.69 & 
        \textbf{0.00} & \textbf{42.05} & \textbf{43.48} & \textbf{33.71} 
        \\   
        R1-SGG (\emph{w/o} cats.) & SFT (\emph{w/o} cats.)  & PSG&
        0.14 & 9.38 & 2.16 & 1.55 &
        0.14 & 41.15 & 41.44 & 31.51 
        \\  
        \bottomrule
    \end{tabular}
    }
    \label{tab:generalization}
\end{table}
We report performance comparisons across datasets in Table~\ref{tab:generalization}. 
The results highlight several key insights:
1) \textbf{VG150 poses a significantly greater challenge than PSG.} 
For instance, SFT trained solely on PSG achieves a high AP@50 of 40.58 and Recall of 24.75\%, with a low failure rate of 0.91\%. In contrast, SFT trained only on VG150 results in a much higher failure rate of 39.54\%, with notably lower AP@50 (14.18) and Recall (9.62\%).
2) \textbf{SFT has a strong domain-specific effect.} 
SFT models trained on one dataset (\eg, VG150) exhibit substantial performance drops when evaluated on another (\eg, PSG), reflecting limited transferability. 
For example, VG150-trained SFT only achieves 3.03\% Recall and 1.36\% mRecall on PSG. 
3) \textbf{Predefined categories in the prompt.}
Models trained and evaluated without categories (denoted as ``\emph{w/o} cats.'') generally exhibit a slight drop in performance,
while those with category information demonstrate better generalization under open-set settings.
4) \textbf{Initialization of RL matters.}
R1-SGG initialized with SFT checkpoints consistently outperforms R1-SGG-Zero. 
On VG150, R1-SGG (7B) achieves 23.75\% Recall and 11.43\% mRecall versus 18.34\% and 8.32\% for R1-SGG-Zero. 
A similar trend is observed on PSG.
This highlights the importance of using SFT as a warm-start for reinforcement learning, which leads to improved sample efficiency and stronger downstream performance.
5) \textbf{R1-SGG-Zero exhibits stronger cross-dataset generalization.} 
This aligns with the domain-specific nature of SFT—models trained via SFT tend to overfit to the source domain, resulting in degraded performance on unseen datasets. 
In contrast, R1-SGG-Zero, trained without SFT, generalizes more robustly across domains.

\begin{table}[t]
    \centering
    \caption{Ablation of reward formulations on VG150 validation set using R1-SGG (7B).}
    \resizebox{0.85\textwidth}{!}{
    \begin{tabular}{lcccccc}
        \toprule
         \textbf{Setting}& \textbf{Sparsity} & \textbf{Metric Aligned} & \textbf{Failure Rate (\%)} & \textbf{AP@50} & \textbf{Recall (\%)} & \textbf{mRecall (\%)} \\
        \midrule
        Hard Recall & sparse & \cmark & 0.08 & 19.47 & 23.75 & 11.43 \\
        Hard Recall + Relax& medium & \xmark & 0.02 & 19.93 & 24.05 & 9.61 \\
        Soft Recall &dense &  \xmark & 0.06 & 18.73 & 21.92 & 5.61 \\
        \bottomrule
    \end{tabular}
    }
    \label{tab:hard_recall}
\end{table}

\noindent\textbf{Hard Recall vs. Soft Recall.} As shown in Table~\ref{tab:hard_recall}, 
\emph{Hard Recall} outperforms other variants despite providing sparser reward signals. 
This highlights the importance of aligning reward functions with evaluation metrics, rather than prioritizing reward smoothness alone.

%%%%%%%%%%%%%%%%%%%%%%%%%%%%%%%%%%%%%%%%%%%%%%%%%%%%%%%%%%%%
\section{Conclusion}
We present a reinforcement learning framework for enhancing end-to-end Scene Graph Generation (SGG) with multimodal large language models (M-LLMs). 
To align training with the structured nature of scene graphs, we design a set of rule-based rewards, comprising three scene graph variants (\emph{Hard Recall}, \emph{Hard Recall+Relax}, and \emph{Soft Recall}) and a format consistency reward, which enable fine-grained and stable policy optimization via GRPO.
Our approach significantly improves the structural validity and relational accuracy of generated scene graphs. 
We release our code and models to support future research on structured visual understanding with M-LLMs.
%%%%%%%%%%%%%%%%%%%%%%%%%%%%%%%%%%%%%%%%%%%%%%%%%%%%%%%%%%%%
\bibliographystyle{named}
\bibliography{main}
%%%%%%%% APPENDIX
\newpage
\appendix
\section{Supplementary Material}
\subsection{Prompt Templates for SGG}
In this work, we adopt two prompt templates for scene graph generation, as illustrated in Table~\ref{tab:prompt_sg} and Table~\ref{tab:prompt_sg_close}.
The difference lies in whether predefined object classes and relation categories are provided.
\begin{table*}[t]
\centering
\caption{Prompting an M-LLM to generate scene graphs without providing predefined object classes or predicate types. 
}
\label{tab:prompt_sg_close}
\definecolor{colorbg@intro}{RGB}{240,240,240}     % Background fill color 
\begin{tikzpicture}[
    node distance=1cm and 0.5cm,
    auto, 
    description/.style={
        rectangle, rounded corners=2pt, draw, thick, fill=colorbg@intro, 
        text width=\textwidth, 
        align=left, %font=\footnotesize,
        scale=1.0
    },
] 
\node[description] {  
\textcolor{blue}{messages} = [\{
    \verb|"|\textbf{role}\verb|"|: \verb|"|system\verb|"|,
    \verb|"|\textbf{content}\verb|"|: \verb|"| \texttt{\{system\_prompt\}}\verb|"|
  \}, 
    \smallskip\{ \verb|"|\textbf{role}\verb|"|: \verb|"|user\verb|"|, \verb|"|\textbf{content}\verb|"|:  f\verb|"""|Generate a structured scene graph for an image using the following format:
```json
{\{
  "objects": [
    {\{"id": "object\_name.number", "bbox": [x1, y1, x2, y2]\}},
    ...
  ],
  "relationships": [
    {\{"subject": "object\_name.number", "predicate": "relationship\_type", "object": "object\_name.number"\}},
    ...
  ]
\}}```.
\verb|###|  **Guidelines:**
- **Objects:**
  - Assign a unique ID for each object using the format "object\_name.number" (e.g., "person.1", "bike.2").
  - Provide its bounding box `[x1, y1, x2, y2]' in integer pixel format.
  - Include all visible objects, even if they have no relationships.

- **Relationships:**
  - Represent interactions accurately using "subject", "predicate", and "object".
  - Omit relationships for orphan objects.

\verb|###| **Example Output:**
```json
\{
  "objects": [
    {\{"id": "person.1", "bbox": [120, 200, 350, 700]\}},
    {\{"id": "bike.2", "bbox": [100, 600, 400, 800]\}},
    {\{"id": "helmet.3", "bbox": [150, 150, 280, 240]\}},
    {\{"id": "tree.4", "bbox": [500, 100, 750, 700]\}}
  ],
  "relationships": [
    {\{"subject": "person.1", "predicate": "riding", "object": "bike.2"\}},
    {\{"subject": "person.1", "predicate": "wearing", "object": "helmet.3"\}}
  ]
\}
```
Now, generate the complete scene graph for the provided image:
\verb|"""|
   \} ]
};
\end{tikzpicture}
\end{table*}
\begin{table*}[t]
\centering
\caption{Prompting an M-LLM to generate scene graphs with predefined object classes and predicate types. 
Here, \emph{OBJ\_CLS} and \emph{REL\_CLS} refer to the predefined object classes and relation categories respectively. 
}
\label{tab:prompt_sg}
\definecolor{colorbg@intro}{RGB}{240,240,240}     % Background fill color 
\begin{tikzpicture}[
    node distance=1cm and 0.5cm,
    auto, 
    description/.style={
        rectangle, rounded corners=2pt, draw, thick, fill=colorbg@intro, 
        text width=\textwidth, 
        align=left, %font=\footnotesize,
        scale=1.0
    },
] 
\node[description] {  
\textcolor{blue}{messages} = [\{
    \verb|"|\textbf{role}\verb|"|: \verb|"|system\verb|"|,
    \verb|"|\textbf{content}\verb|"|: \verb|"| \texttt{\{system\_prompt\}}\verb|"|
  \}, 
    \smallskip\{ \verb|"|\textbf{role}\verb|"|: \verb|"|user\verb|"|, \verb|"|\textbf{content}\verb|"|:  f\verb|"""|Generate a structured scene graph for an image using the following format:
```json
{\{
  "objects": [
    {\{"id": "object\_name.number", "bbox": [x1, y1, x2, y2]\}},
    ...
  ],
  "relationships": [
    {\{"subject": "object\_name.number", "predicate": "relationship\_type", "object": "object\_name.number"\}},
    ...
  ]
\}}```.
\verb|###|  **Guidelines:**
- **Objects:**
  - Assign a unique ID for each object using the format "object\_name.number" (e.g., "person.1", "bike.2").
  \textcolor{red}{The **object\_name** must belong to the predefined object set: `\{OBJ\_CLS\}'}.
  - Provide its bounding box `[x1, y1, x2, y2]' in integer pixel format.
  - Include all visible objects, even if they have no relationships.

- **Relationships:**
  - Represent interactions accurately using "subject", "predicate", and "object".
  - Omit relationships for orphan objects.
  - \textcolor{red}{The **predicate** must belong to the predefined relationship set: `\{REL\_CLS\}'}.
\verb|###| **Example Output:**
```json
\{
  "objects": [
    {\{"id": "person.1", "bbox": [120, 200, 350, 700]\}},
    {\{"id": "bike.2", "bbox": [100, 600, 400, 800]\}},
    {\{"id": "helmet.3", "bbox": [150, 150, 280, 240]\}},
    {\{"id": "tree.4", "bbox": [500, 100, 750, 700]\}}
  ],
  "relationships": [
    {\{"subject": "person.1", "predicate": "riding", "object": "bike.2"\}},
    {\{"subject": "person.1", "predicate": "wearing", "object": "helmet.3"\}}
  ]
\}
```
Now, generate the complete scene graph for the provided image:
\verb|"""|
   \} ]
};
\end{tikzpicture}
\end{table*}

\subsection{How Well Do M-LLMs Reason About Visual Relationships?}
\begin{table}[t]
\centering
\caption{Comparison of VQA on the VG150 validation set across various models and settings. Gains compared to the \textit{Original Image} (1st row) are indicated in red.
\textit{``mask img.''} refers to masking the entire image with random noise, 
\textit{``mask obj.''} refers to masking object regions with black pixels, 
\textit{``w/o cats.''} refers to not providing object categories in the prompt, and 
\textit{``w/o box.''} refers to not providing bounding boxes in the prompt.
}
\resizebox{\textwidth}{!}{%
\begin{tabular}{lcccccccc}
\toprule
 & \multicolumn{2}{c}{\textbf{InstructBLIP 7B}} & \multicolumn{2}{c}{\textbf{LLaVA v1.5 7B}} & \multicolumn{2}{c}{\textbf{LLaVA v1.6 7B}} & \multicolumn{2}{c}{\textbf{Qwen2VL 7B}} \\
\cmidrule(lr){2-3} \cmidrule(lr){4-5} \cmidrule(lr){6-7} \cmidrule(lr){8-9} 
& \textbf{Acc} & \textbf{mAcc} & \textbf{Acc} & \textbf{mAcc} & \textbf{Acc} & \textbf{mAcc} & \textbf{Acc} & \textbf{mAcc} \\
\midrule
org. img. &  2.3 & 1.9 & 45.8 & 45.6 & 28.7 & 29.2 & 53.7 & 53.4 \\
mask img.& 1.0 \textcolor{red}{\scriptsize(-1.3)} & 1.0 \textcolor{red}{\scriptsize(-0.9)} & 21.8 \textcolor{red}{\scriptsize(-24.0)} & 21.6 \textcolor{red}{\scriptsize(-24.0)} & 3.9 \textcolor{red}{\scriptsize(-24.8)} & 4.0 \textcolor{red}{\scriptsize(-25.2)} & 0.0 \textcolor{red}{\scriptsize(-53.7)} & 0.0 \textcolor{red}{\scriptsize(-53.4)} \\
mask obj. & 1.9 \textcolor{red}{\scriptsize(-0.4)} & 1.9 \textcolor{red}{\scriptsize(-0.1)} & 37.2 \textcolor{red}{\scriptsize(-8.6)} & 37.2 \textcolor{red}{\scriptsize(-8.4)} & 12.8 \textcolor{red}{\scriptsize(-15.9)} & 13.2 \textcolor{red}{\scriptsize(-16.0)} & 16.2 \textcolor{red}{\scriptsize(-37.5)} & 16.8 \textcolor{red}{\scriptsize(-36.5)} \\
\midrule
w/o cats.&  2.5 \textcolor{red}{\scriptsize(+0.2)} & 2.4 \textcolor{red}{\scriptsize(+0.4)} & 32.8 \textcolor{red}{\scriptsize(-12.9)} & 32.7 \textcolor{red}{\scriptsize(-12.9)} & 9.5 \textcolor{red}{\scriptsize(-19.2)} & 10.1 \textcolor{red}{\scriptsize(-19.1)} & 16.8 \textcolor{red}{\scriptsize(-36.9)} & 18.1 \textcolor{red}{\scriptsize(-35.3)} \\
\quad+ mask img.&  1.0 \textcolor{red}{\scriptsize(-1.3)} & 1.0 \textcolor{red}{\scriptsize(-0.9)} & 15.4 \textcolor{red}{\scriptsize(-30.3)} & 15.3 \textcolor{red}{\scriptsize(-30.3)} & 0.0 \textcolor{red}{\scriptsize(-28.7)} & 0.0 \textcolor{red}{\scriptsize(-29.2)} & 0.2 \textcolor{red}{\scriptsize(-53.6)} & 0.2 \textcolor{red}{\scriptsize(-53.1)} \\
\quad+ mask obj.&  1.8 \textcolor{red}{\scriptsize(-0.5)} & 1.7 \textcolor{red}{\scriptsize(-0.3)} & 27.9 \textcolor{red}{\scriptsize(-17.8)} & 28.4 \textcolor{red}{\scriptsize(-17.2)} & 3.3 \textcolor{red}{\scriptsize(-25.4)} & 3.8 \textcolor{red}{\scriptsize(-25.4)} & 4.7 \textcolor{red}{\scriptsize(-49.1)} & 5.5 \textcolor{red}{\scriptsize(-47.8)} \\
\midrule
w/o box. & 26.0 \textcolor{red}{\scriptsize(+23.7)} & 25.9 \textcolor{red}{\scriptsize(+24.0)} & 61.9 \textcolor{red}{\scriptsize(+16.2)} & 61.3 \textcolor{red}{\scriptsize(+15.7)} & 53.5 \textcolor{red}{\scriptsize(+24.8)} & 52.1 \textcolor{red}{\scriptsize(+22.9)} & 78.1 \textcolor{red}{\scriptsize(+24.4)} & 77.1 \textcolor{red}{\scriptsize(+23.8)} \\
\quad+ mask img.& 10.1 \textcolor{red}{\scriptsize(+7.9)} & 10.2 \textcolor{red}{\scriptsize(+8.2)} & 36.3 \textcolor{red}{\scriptsize(-9.5)} & 35.2 \textcolor{red}{\scriptsize(-10.4)} & 11.5 \textcolor{red}{\scriptsize(-17.2)} & 11.4 \textcolor{red}{\scriptsize(-17.7)} & 0.0 \textcolor{red}{\scriptsize(-53.7)} & 0.0 \textcolor{red}{\scriptsize(-53.4)} \\
\quad+ mask obj.& 19.3 \textcolor{red}{\scriptsize(+17.0)} & 19.1 \textcolor{red}{\scriptsize(+17.1)} & 54.2 \textcolor{red}{\scriptsize(+8.5)} & 53.8 \textcolor{red}{\scriptsize(+8.2)} & 33.5 \textcolor{red}{\scriptsize(+4.8)} & 33.2 \textcolor{red}{\scriptsize(+4.1)} & 40.3 \textcolor{red}{\scriptsize(-13.4)} & 39.3 \textcolor{red}{\scriptsize(-14.1)} \\
\bottomrule
\end{tabular}%
}
\label{tab:accuracy_comparison}
\end{table}
To evaluate the reasoning capabilities of M-LLMs over visual relationships, 
we present results in Table~\ref{tab:accuracy_comparison}.
We vary both the visual input and the text prompt conditions to assess robustness.
For visual variations, we consider: \emph{org. img.}, \emph{mask img.}, and \emph{mask obj.};
for prompt variations, we add: \emph{w/o cats.} (without object categories) and \emph{w/o box.} (without bounding boxes).

%%%%%%%%%%%%%%%%%%%%%%%
\begin{figure*}[htbp]
\centering
% SFT
\definecolor{color2b}{HTML}{0E6E57} % 深绿色
\definecolor{color7b}{HTML}{6E0E25} % 深红色

\definecolor{c2}{HTML}{1F78B4}  % R1-SGG-Zero (2B) - Blue
\definecolor{c4}{HTML}{FF7F0E}  % R1-SGG-Zero (7B) - Orange
\definecolor{c6}{HTML}{6A3D9A}  % R1-SGG (2B) - Purple
\definecolor{c8}{HTML}{E31A1C}  % R1-SGG (7B) - Red

\resizebox{\textwidth}{!}{
\begin{tikzpicture}
\begin{groupplot}[
    group style={
        group size=3 by 2,
        vertical sep=2cm,
        horizontal sep=1cm
    },
    width=5.2cm,
    height=4.2cm,
    axis line style={gray},
    grid=both,
    grid style={dashed, gray!30},
    tick label style={font=\scriptsize},
    label style={font=\scriptsize},
    xlabel={Training Steps},
    xlabel near ticks,
    ylabel near ticks,
    xtick={0,400,800,1200,1600},
]
% First Row --- SFT Plots ---
% Failure Rate
\nextgroupplot[
    ylabel={Failure Rate (\%)},
    ymin=20, ymax=110,
    mark size=1pt,
    line width=0.8pt,
    xlabel style={at={(axis description cs:0.5,-0.13)}, anchor=north, font=\scriptsize},
    ylabel style={at={(axis description cs:-0.1,0.5)}, anchor=south, font=\scriptsize},
]
\addplot+[mark=o, color=color2b, thick] coordinates {
  (0, 59.96) (100, 99.84) (200, 93.6) (300, 81.26) (400, 82.94) (500, 90.86) (600, 82.28) (700, 70.94) (800, 75.9) (900, 78.9) (1000, 76.12) (1100, 73.66) (1200, 74.5) (1300, 72.6)
};
\addplot+[mark=square*, color=color7b, thick] coordinates {
    (0, 54.46)  (100, 94.58) (200, 83.46) (300, 43.26) (400, 55.02) (500, 55.14) (600, 52.16) (700, 61.04) (800, 40.34) (900, 31.02) (1000, 37.8) (1100, 32.28) (1200, 39.26) (1300, 39.14)
};

% AP@50
\nextgroupplot[
    ylabel={AP@50},
    ymin=-1, ymax=16,
    mark size=1pt,
    line width=0.8pt,
    xlabel style={at={(axis description cs:0.5,-0.13)}, anchor=north, font=\scriptsize},
    ylabel style={at={(axis description cs:-0.1,0.5)}, anchor=south, font=\scriptsize},
]
\addplot+[mark=o, color=color2b, thick] coordinates {
    (0, 2.18) (100, 0.059) (200, 2.668) (300, 6.005) (400, 5.473) (500, 3.695) (600, 6.06) (700, 8.528) (800, 6.992) (900, 6.577) (1000, 7.357) (1100, 7.993) (1200, 7.616) (1300, 8.486)
};
\addplot+[mark=square*, color=color7b, thick] coordinates {
    (0, 6.07) (100, 2.294) (200, 5.594) (300, 13.007) (400, 11.185) (500, 11.397) (600, 11.969) (700, 10.886) (800, 13.842) (900, 14.241) (1000, 14.586) (1100, 14.901) (1200, 14.072) (1300, 14.65)
};

% Recall
\nextgroupplot[
    ylabel={Recall (\%)},
    ymin=-1, ymax=12,
    mark size=1pt,
    line width=0.8pt,
    xlabel style={at={(axis description cs:0.5,-0.13)}, anchor=north, font=\scriptsize},
    ylabel style={at={(axis description cs:-0.1,0.5)}, anchor=south, font=\scriptsize},
]
\addplot+[mark=o, color=color2b, thick] coordinates {
    (0, 0.07)  (100, 0.07) (200, 1.31) (300, 3.53) (400, 3.54) (500, 2.29) (600, 3.77) (700, 4.84) (800, 4.52) (900, 4.35) (1000, 4.77) (1100, 5.13) (1200, 5.14) (1300, 5.21)
};
\addplot+[mark=square*, color=color7b, thick] coordinates {
    (0,0.69)    (100, 1.29) (200, 3.3) (300, 7.74) (400, 7.39) (500, 7.47) (600, 8.34) (700, 7.01) (800, 9.88) (900, 10.36) (1000, 9.91) (1100, 11) (1200, 10.11) (1300, 10.21)
};

%%%%%%%%%%%%%%%%%%%%%%%%%%%%%%%%%%
% --- Failure Rate Plot ---
\nextgroupplot[
    ylabel={Failure Rate (\%)},
    ymin=-10, ymax=100,
    mark size=1pt,
    line width=0.8pt,
    xlabel style={at={(axis description cs:0.5,-0.13)}, anchor=north, font=\scriptsize},
    ylabel style={at={(axis description cs:-0.1,0.5)}, anchor=south, font=\scriptsize},
]
% R1-SGG-Zero, 2B
\addplot+[mark=triangle, color=c2, thick] coordinates {
(0, 59.96) (100, 21.42) (200, 70.14) (300, 47.2) (400, 58.02) (500, 9.4) (600, 4.68) (700, 0.36) (800, 1.02) (900, 0.92) (1000, 0.42) (1100, 0.42) (1200, 0.06) (1300, 0.24) (1400, 0.2) (1500, 0.34) (1600, 0.38) (1700, 0.44)
};
% R1-SGG-Zero, 7B
\addplot+[mark=diamond, color=c4, thick] coordinates {
(0, 54.46)   (100, 12.36) (200, 1.12) (300, 0.26) (400, 0.08) (500, 0.06) (600, 0.5) (700, 0.04) (800, 0.28) (900, 0.1) (1000, 0.26) (1100, 0.26) (1200, 0.14) (1300, 0.06) (1400, 0.04) (1500, 0.06) (1600, 0.02) (1700, 0.02)  
};
% R1-SGG,2B
\addplot+[mark=triangle, color=c6, thick] coordinates {
(0, 72.42)    (100, 74.7) (200, 20.62) (300, 6.5) (400, 1.26) (500, 1.1) (600, 0.24) (700, 0.6) (800, 0.32) (900, 0.4) (1000, 0.18) (1100, 0.1) (1200, 0.16) (1300, 0.1) (1400, 0.02) (1500, 0.1) (1600, 0.1) (1700, 0.08)  
};
% R1-SGG, 7B
\addplot+[mark=diamond, color=c8, thick] coordinates {
(0, 39.54)   (100, 50.48) (200, 20.46) (300, 2.72) (400, 0.46) (500, 0.22) (600, 0.1) (700, 0.18) (800, 0.34) (900, 0.2) (1000, 0.46) (1100, 0.12) (1200, 0.12) (1300, 0.08) (1400, 0.02) (1500, 0.12) (1600, 0.14) (1700, 0.08)
};

% --- AP@50 Plot ---
\nextgroupplot[
    ylabel={AP@50},
    ymin=0, ymax=21,
    mark size=1pt,
    line width=0.8pt,
    xlabel style={at={(axis description cs:0.5,-0.13)}, anchor=north, font=\scriptsize},
    ylabel style={at={(axis description cs:-0.1,0.5)}, anchor=south, font=\scriptsize},    
]

\addplot+[mark=triangle, color=c2, thick] coordinates {
(0, 2.18)  (100, 5.222) (200, 4.31) (300, 5.928) (400, 5.422) (500, 9.795) (600, 10.703) (700, 11.169) (800, 11.249) (900, 11.884) (1000, 11.62) (1100, 11.807) (1200, 12.045) (1300, 12.099) (1400, 12.202) (1500, 12.493) (1600, 12.23) (1700, 12.177)  
};

\addplot+[mark=diamond, color=c4, thick] coordinates {
(0,6.07)    (100, 9.544) (200, 12.616) (300, 14.632) (400, 14.456) (500, 14.736) (600, 14.906) (700, 14.921) (800, 15.423) (900, 14.726) (1000, 14.768) (1100, 15.262) (1200, 15.673) (1300, 15.413) (1400, 15.457) (1500, 15.583) (1600, 15.478) (1700, 15.552)
};
 
\addplot+[mark=triangle, color=c6, thick] coordinates {
(0,  8.10)    (100, 6.5) (200, 14.641) (300, 16.847) (400, 17.246) (500, 17.345) (600, 17.811) (700, 17.525) (800, 17.537) (900, 17.669) (1000, 17.75) (1100, 17.634) (1200, 17.785) (1300, 17.787) (1400, 17.816) (1500, 17.872) (1600, 17.953) (1700, 18.075)  
};
 
\addplot+[mark=diamond, color=c8, thick] coordinates {
(0, 14.18)   (100, 11.514) (200, 16.105) (300, 18.19) (400, 18.626) (500, 18.889) (600, 19.063) (700, 19.052) (800, 18.999) (900, 19.206) (1000, 19.292) (1100, 19.389) (1200, 19.356) (1300, 19.411) (1400, 19.52) (1500, 19.383) (1600, 19.469) (1700, 19.52)  
}; 

% --- Recall Plot ---
\nextgroupplot[
    ylabel={Recall (\%)},
    ymin=-1, ymax=25,
    mark size=1pt,
    line width=0.8pt,
    xlabel style={at={(axis description cs:0.5,-0.13)}, anchor=north, font=\scriptsize},
    ylabel style={at={(axis description cs:-0.1,0.5)}, anchor=south, font=\scriptsize},    
]

\addplot+[mark=triangle, color=c2, thick] coordinates {
    (0,0.07)    (100, 0.53) (200, 1.33) (300, 3.4) (400, 3.18) (500, 7.15) (600, 8.31) (700, 8.31) (800, 9.26) (900, 9.18) (1000, 10.22) (1100, 10.49) (1200, 10.6) (1300, 11.16) (1400, 11.31) (1500, 11.79) (1600, 11.78) (1700, 11.89) 
};

\addplot+[mark=diamond, color=c4, thick] coordinates {
    (0,0.69)    (100, 6.62) (200, 10.27) (300, 13.57) (400, 13.97) (500, 14.69) (600, 14.48) (700, 15.53) (800, 16.21) (900, 16.8) (1000, 16.94) (1100, 17.55) (1200, 17.57) (1300, 17.21) (1400, 17.81) (1500, 18.02) (1600, 18.15) (1700, 18.41)  
};
 
\addplot+[mark=triangle, color=c6, thick] coordinates {
(0,  5.47)    (100, 5.77) (200, 14.27) (300, 17.86) (400, 18.72) (500, 19.42) (600, 18.43) (700, 19.22) (800, 19.59) (900, 19.5) (1000, 19.33) (1100, 20.04) (1200, 20.76) (1300, 20.75) (1400, 20.93) (1500, 21.04) (1600, 20.87) (1700, 21.05)  
};

\addplot+[mark=diamond, color=c8, thick] coordinates {
(0, 9.62)   (100, 10.49) (200, 16.65) (300, 19.27) (400, 20.28) (500, 20.18) (600, 20.54) (700, 21.23) (800, 21.72) (900, 21.93) (1000, 22.74) (1100, 22.61) (1200, 23.32) (1300, 23.54) (1400, 23.7) (1500, 23.8) (1600, 23.56) (1700, 23.72)  
}; 
 
\end{groupplot}
%%%%%%%%%%%%%%% legend 
% Unified Legend (Top - for SFT row)
\node at ($(group c1r1.north)!0.5!(group c3r1.north) + (0, 1em)$) {%
\begin{tabular}{c c}
\tikz{\draw[color2b, line width=0.8pt] (0,0) -- (0.7,0) node[pos=0.5, circle, draw=color2b, fill=white, inner sep=1pt] {};} \, {\scriptsize Qwen2-VL-2B-Instruct (SFT)}
\hspace{1cm}
& 
\tikz{\draw[color7b, line width=0.8pt] (0,0) -- (0.7,0) node[pos=0.5, rectangle, draw=color7b, fill=white, inner sep=1pt] {};} \, {\scriptsize Qwen2-VL-7B-Instruct (SFT)}
\end{tabular}
}; 

\node at ($(group c1r2.north)!0.5!(group c3r2.north) + (0,2em)$) {
\begin{tabular}{@{\hskip 5pt}l@{\hskip 5pt}l@{\hskip 5pt}l@{\hskip 5pt}l}
\tikz{\draw[c2, thick] (0,0) -- (0.5,0) node[pos=0.5, regular polygon, regular polygon sides=3, draw=c2, fill=white, inner sep=1pt] {};} {\scriptsize R1-SGG-Zero (2B)} &
\tikz{\draw[c4, thick] (0,0) -- (0.5,0) node[pos=0.5, diamond, draw=c4, fill=white, inner sep=1pt] {};} {\scriptsize R1-SGG-Zero (7B)} &
\tikz{\draw[c6, thick] (0,0) -- (0.5,0) node[pos=0.5, regular polygon, regular polygon sides=3, draw=c6, fill=white, inner sep=1pt] {};} {\scriptsize R1-SGG (2B)} &
\tikz{\draw[c8, thick] (0,0) -- (0.5,0) node[pos=0.5, diamond, draw=c8, fill=white, inner sep=1pt] {};} {\scriptsize R1-SGG (7B)}
\end{tabular}
};
%%%%%%%%%%%
\end{tikzpicture}
}
\caption{
Comparison of  {R1-SGG-Zero} and  {R1-SGG} models against {SFT baselines} (Qwen2-VL-2B/7B-Instruct) across training steps on the VG150 validation set in terms of {Failure Rate} (\%), {AP@50}, and  {Recall} (\%).
}
\label{fig:zero_rl_metrics_steps}
\end{figure*}
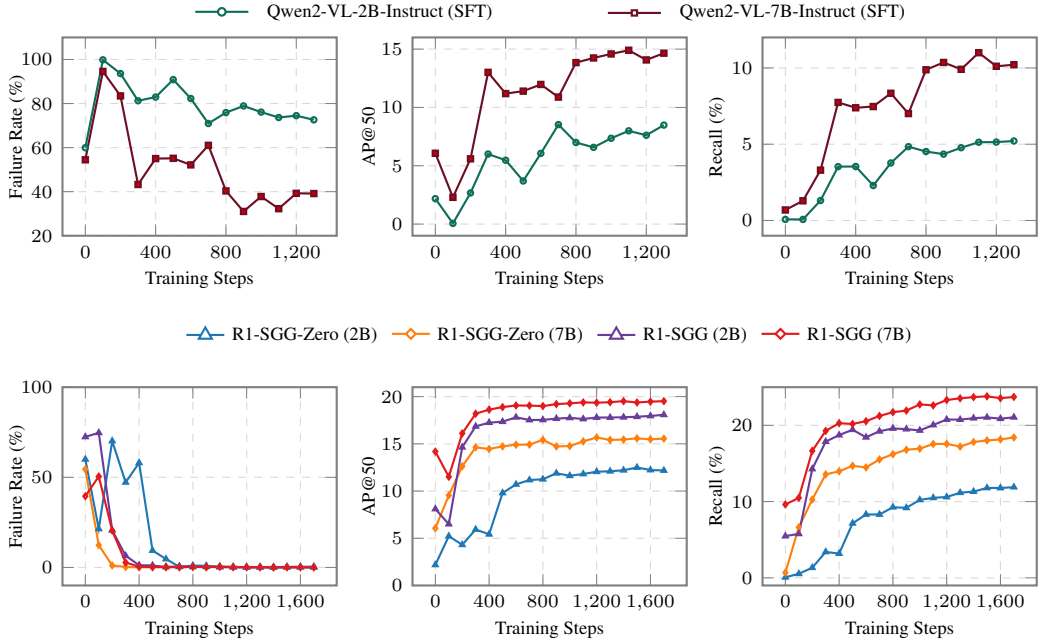

\subsection{Qualitative Results}
\begin{figure}[htbp]
    \centering
    \resizebox{\textwidth}{0.88\textheight}
    {
    \begin{tikzpicture}[node distance=1cm]
        \node (gt) at (0, 0) {
            \includegraphics[width=0.3\textwidth]{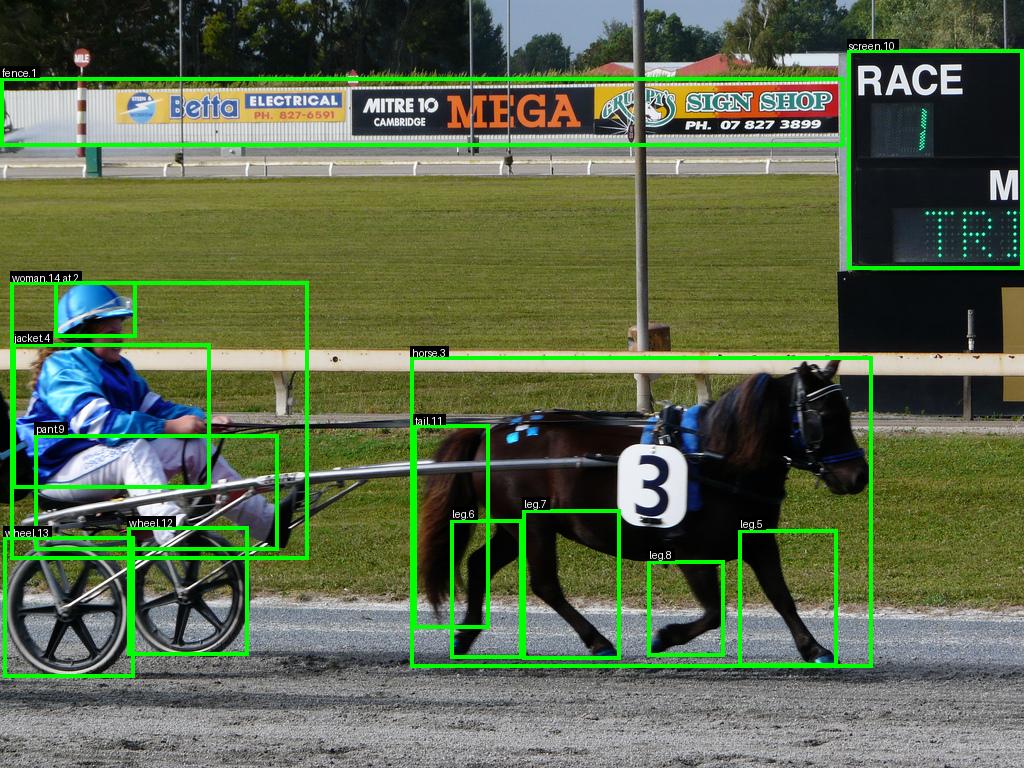}
        };
        \node (gt_sg) [right=of gt, xshift=-8mm]{
            \includegraphics[width=0.4\textwidth]{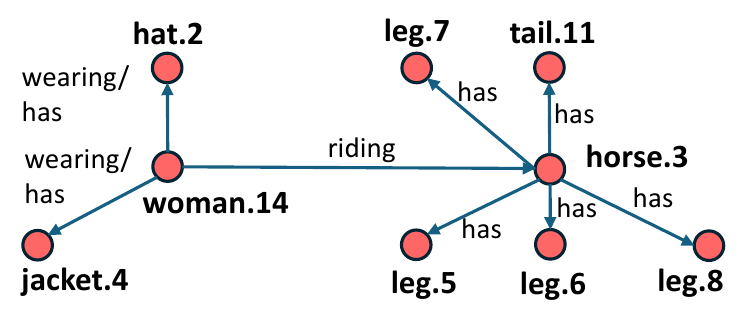}
        };
        \node [below=of gt_sg, yshift=12mm] {(a) GT.};
    % row 1
        \node (a00) [below=of gt, yshift=10mm] {
            \includegraphics[width=0.3\textwidth]{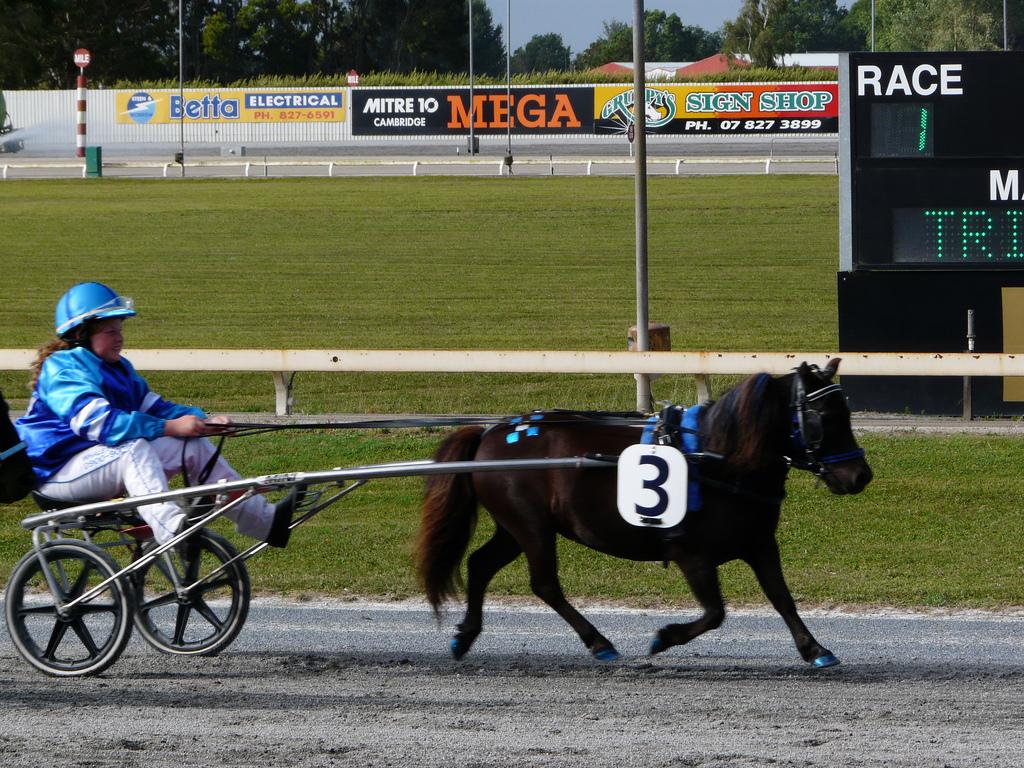}
        };
         \node (a01) [right=of a00, xshift=-5mm] {
            \includegraphics[width=0.35\textwidth]{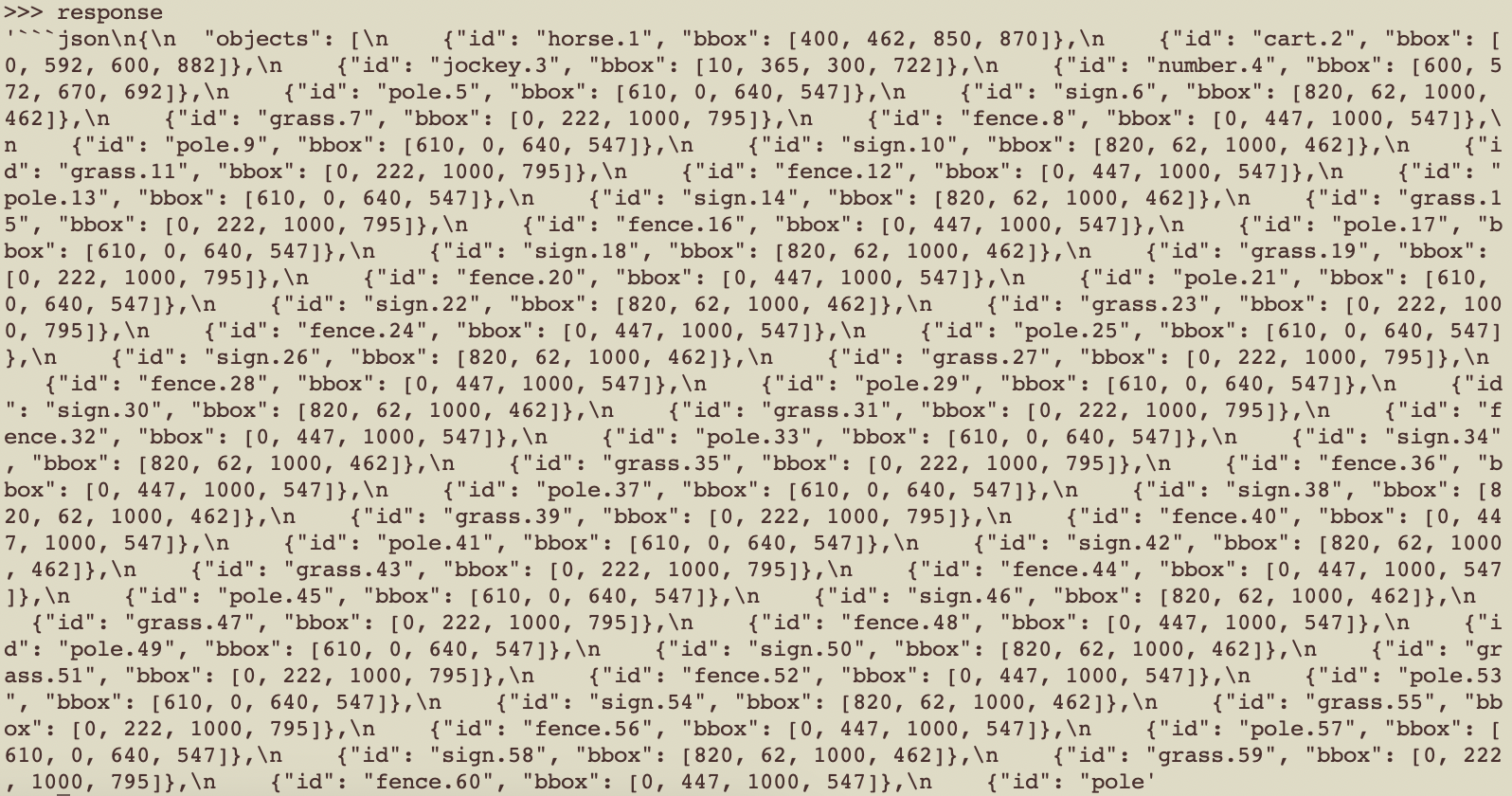}
        };    
      %  \node (a02) [above=of a01, yshift=-12mm] {Invalid Response};
        \node [below=of a01, yshift=12mm] {(b) Qwen2-VL-7B-Instruct.};
    % row 2
        \node (a10) [below =of a00, yshift=10mm] {
            \includegraphics[width=0.3\textwidth]{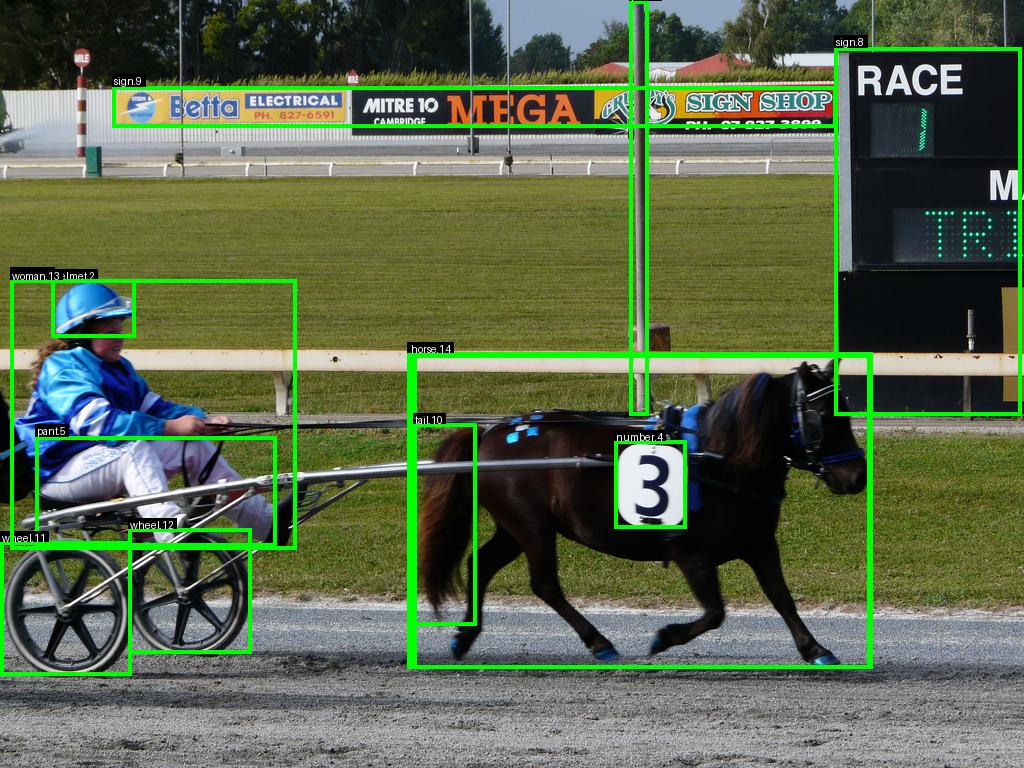}
        }; 
         % scene graph 
        \node (a11) [right=of a10, xshift=-8mm] {
            \includegraphics[width=0.45\textwidth]{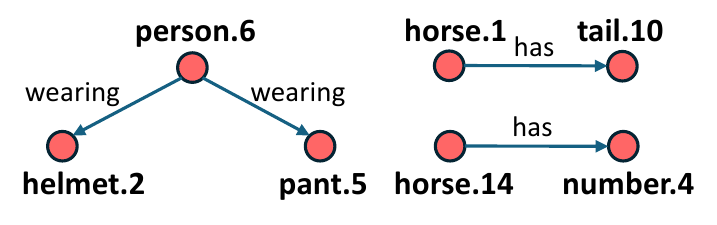}
        };
         \node [below=of a11, yshift=12mm] {(c) Qwen2-VL-7B-Instruct (SFT).};
   % row 3
        \node (a20) [below =of a10, yshift=10mm] {
            \includegraphics[width=0.3\textwidth]{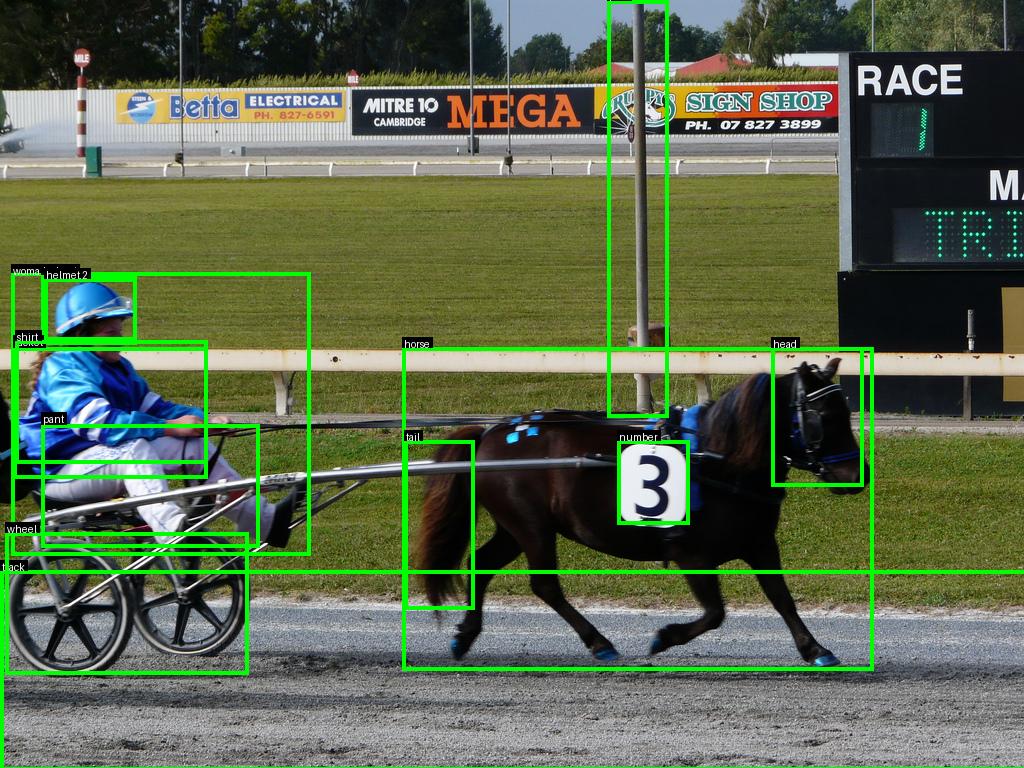}
        }; 
        % scene graph 
        \node (a21) [right=of a20, xshift=-8mm] {
            \includegraphics[width=0.45\textwidth]{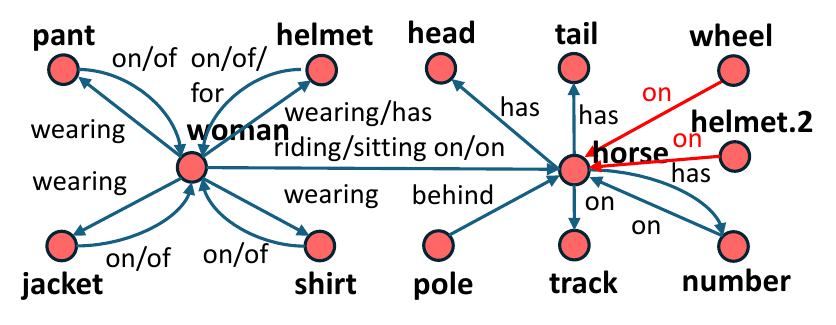}
        };
        \node [below=of a21, yshift=12mm] {(d) R1-SGG-Zero (7B).};
   % row 4 
        \node (a30) [below =of a20, yshift=10mm] {
            \includegraphics[width=0.3\textwidth]{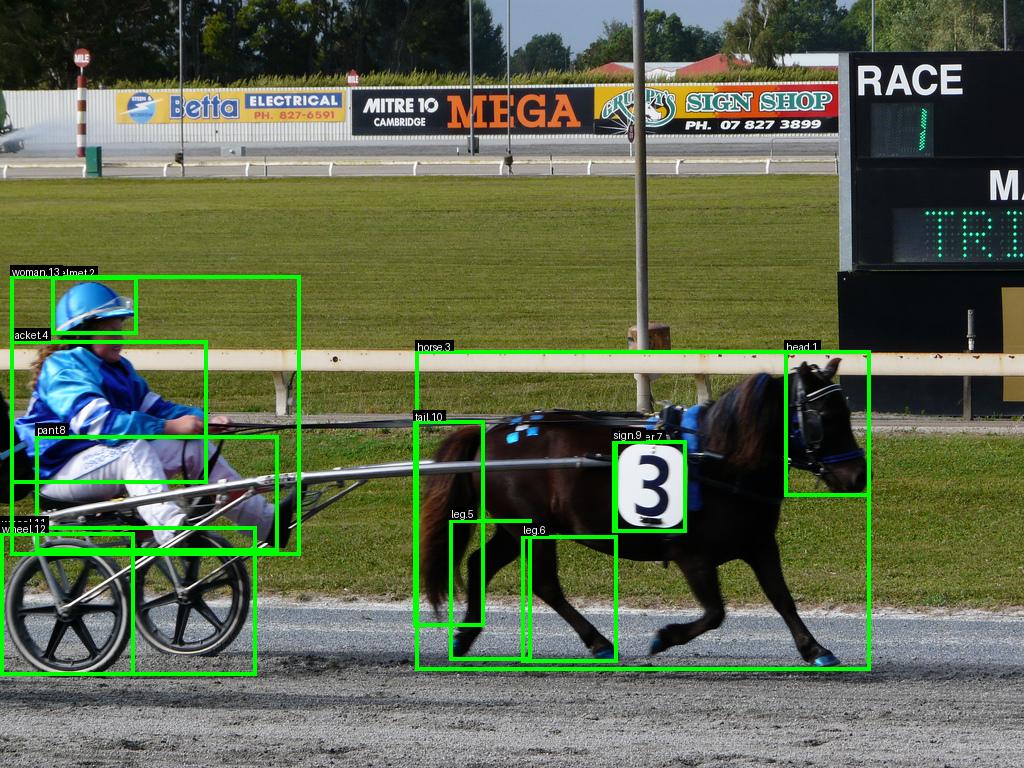}
        }; 
        % scene graph 
        \node (a31) [right=of a30, xshift=-8mm]{
            \includegraphics[width=0.45\textwidth]{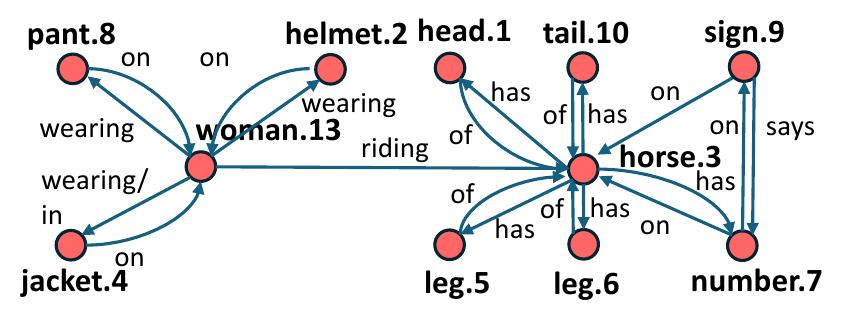}
        }; 
        \node [below=of a31, yshift=12mm]{(e) R1-SGG (7B).};
          
    \end{tikzpicture}
} 
\caption{Qualitative comparison of generated scene graphs (from VG150). 
(a) Ground-truth scene graph annotated by humans. 
(b) Zero-shot {Qwen2-VL-7B-Instruct} produces an invalid JSON (failure to follow format). 
(c) Qwen2-VL-7B-Instruct (SFT) outputs a valid graph but omits many relations. 
(d) R1-SGG-Zero (7B) recovers most objects and relations but still hallucinates incorrect triplets (\eg, \emph{$<$ wheel, on, horse$>$}  and \emph{$<$ helmet.2, on, horse $>$}). 
(e) R1-SGG (7B) yields a complete, structurally correct scene graph with higher recall.
}
    \label{fig:qual_vis_vg}
\end{figure}

\begin{figure}[htbp]
    \centering
    \resizebox{\textwidth}{0.88\textheight}
    {
    \begin{tikzpicture}[node distance=1cm]
        \node (gt) at (0, 0) {
            \includegraphics[width=0.3\textwidth]{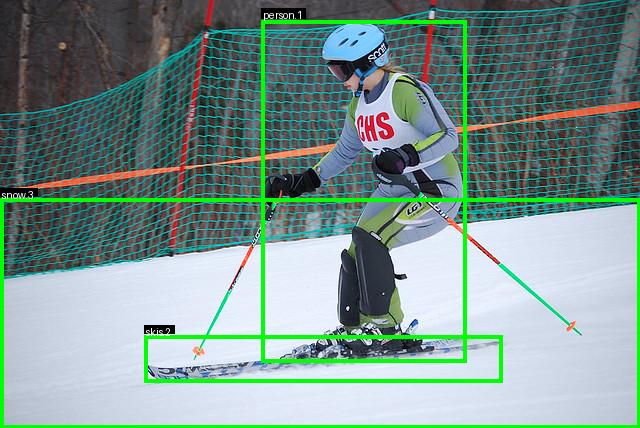}
        };
        \node (gt_sg) [right=of gt, xshift=0mm]{
            \includegraphics[width=0.25\textwidth]{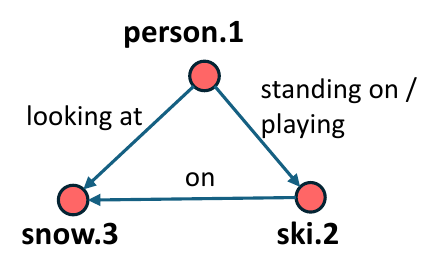}
        };
        \node [below=of gt_sg, yshift=12mm] {(a) GT.};
    % row 1
        \node (a00) [below=of gt, yshift=10mm] {
            \includegraphics[width=0.3\textwidth]{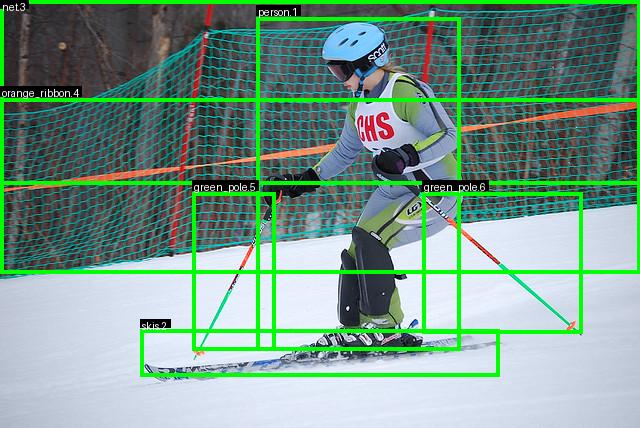}
        };
         \node (a01) [right=of a00, xshift=0mm, yshift=2mm] {
            \includegraphics[width=0.25\textwidth]{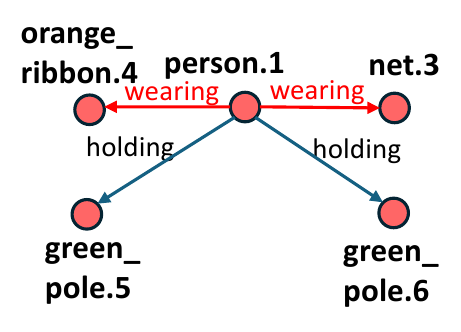}
        };    
        \node [below=of a01, yshift=12mm] {(b) Qwen2-VL-7B-Instruct.};
    % row 2
        \node (a10) [below =of a00, yshift=10mm] {
            \includegraphics[width=0.3\textwidth]{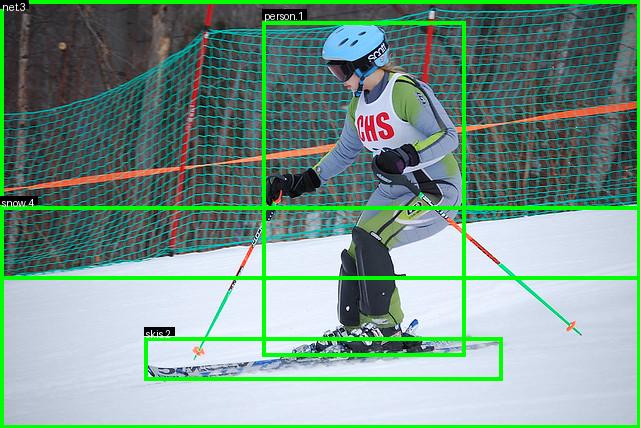}
        }; 
         % scene graph 
        \node (a11) [right=of a10, xshift=0mm, yshift=2mm] {
            \includegraphics[width=0.25\textwidth]{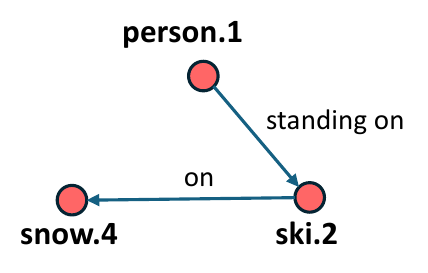}
        };
         \node [below=of a11, yshift=12mm] {(c) Qwen2-VL-7B-Instruct (SFT).};
   % row 3
        \node (a20) [below =of a10, yshift=10mm] {
            \includegraphics[width=0.3\textwidth]{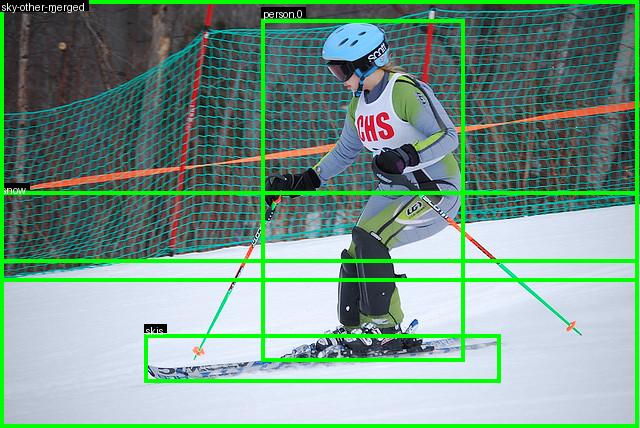}
        }; 
        % scene graph 
        \node (a21) [right=of a20, xshift=-5mm, yshift=2mm] {
            \includegraphics[width=0.35\textwidth]{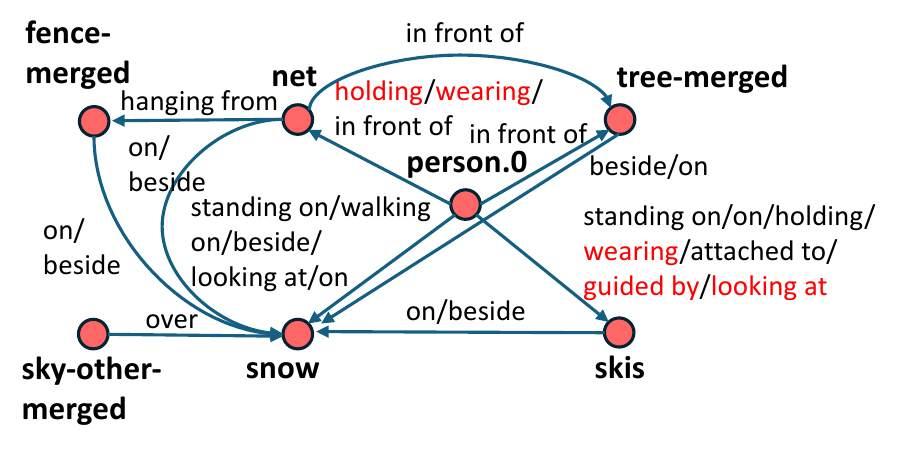}
        };
        \node [below=of a21, yshift=12mm] {(d) R1-SGG-Zero (7B).};
   % row 4 
        \node (a30) [below =of a20, yshift=10mm] {
            \includegraphics[width=0.3\textwidth]{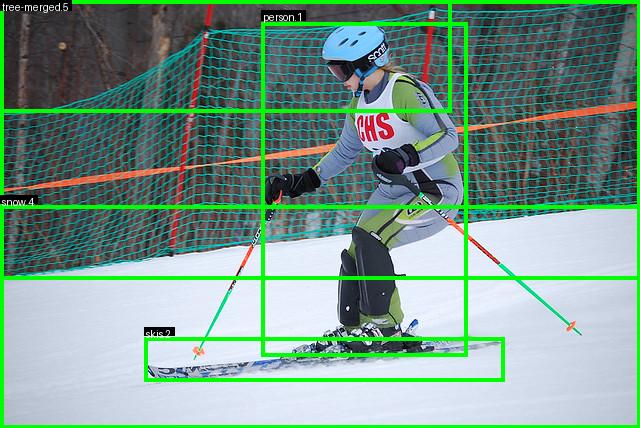}
        }; 
        % scene graph 

        \node (a31) [right=of a30, xshift=-5mm, yshift=2mm]{
            \includegraphics[width=0.35\textwidth]{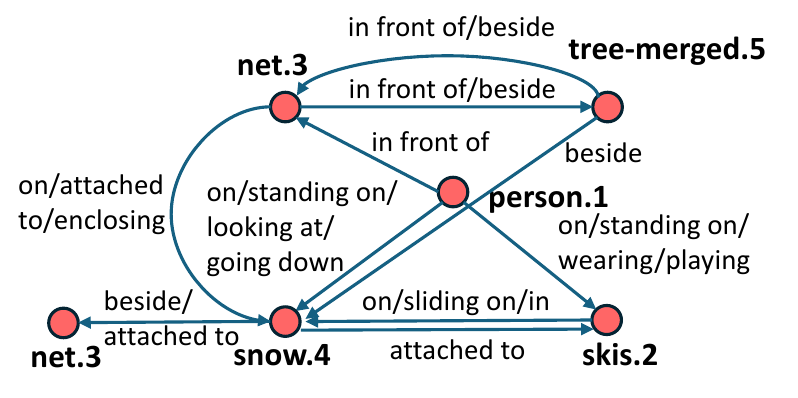}
        }; 
        \node [below=of a31, yshift=12mm]{(e) R1-SGG (7B).};
          
    \end{tikzpicture}
} 
\caption{Qualitative comparison of generated scene graphs (from PSG). 
(a) Ground-truth scene graph annotated by humans. 
(b) Zero-shot {Qwen2-VL-7B-Instruct} generates a valid graph but includes incorrect triplets (\eg, \emph{$<$person.1, wearing, net.3$>$}). 
(c) {Qwen2-VL-7B-Instruct (SFT)} produces a valid graph but omits some relationships. 
(d) {R1-SGG-Zero (7B)} recovers most objects and relations but still hallucinates errors (\eg, \emph{$<$person.0, wearing, net$>$}). 
(e) {R1-SGG (7B)}  generates a complete and accurate scene graph with higher recall.
}
    \label{fig:qual_vis_psg}
\end{figure}

\input{pics/recall_head_tails}
We present qualitative results in Fig.~\ref{fig:qual_vis_vg} and Fig.~\ref{fig:qual_vis_psg}, and analyze head and tail predicate performance in Fig.~\ref{fig:recall_head_tails} and Fig.~\ref{fig:recall_head_tails_psg} to assess long-tail bias.
As shown in Fig.~\ref{fig:recall_head_tails}, both specific models such as OvSGTR and M-LLMs like Qwen2-VL-7B-Instruct (with or without SFT) tend to be biased toward head classes, whereas R1-SGG achieves significantly higher recall on tail predicates.
This trend is also confirmed on the PSG dataset in Fig.~\ref{fig:recall_head_tails_psg}.
These results demonstrate that R1-SGG is more effective at generating unbiased scene graphs.

\begin{figure*}[htbp]
\centering

% --- Color Definitions ---
%\definecolor{color2b_a}{HTML}{0E6E57} % 深绿色
%\definecolor{color2b_b}{HTML}{6E0E25} % 深红色

% Row 1: KL vs. no KL
\definecolor{kl_green}{HTML}{0E6E57}    % Deep Green
\definecolor{kl_red}{HTML}{6E0E25}      % Deep Red

% Row 2: Sampling length
\definecolor{sample_blue}{HTML}{0E6E57} % 
\definecolor{sample_purple}{HTML}{6E0E25} %

% Row 3: Group size
\definecolor{group_orange}{HTML}{0E6E57}  
\definecolor{group_brown}{HTML}{6E0E25}   

\resizebox{\textwidth}{!}{
\begin{tikzpicture}
\begin{groupplot}[
    group style={
        group size=3 by 3, 
        vertical sep=2cm,
        horizontal sep=1cm,
    },
    width=5.2cm,
    height=4.2cm,
    axis line style={gray},
    grid=both,
    grid style={dashed, gray!30},
    tick label style={font=\scriptsize},
    label style={font=\scriptsize},
    xlabel={Training Steps},
    xlabel near ticks,
    ylabel near ticks,
    xtick={0,400,800,1200, 1600},
]
%%%%%%%%%%%%% KL vs. no KL %%%%%%%%%%%%%%%%
% --- Failure Rate Plot ---
\nextgroupplot[
    ylabel={Failure Rate (\%)},
    ymin=-5, ymax=80,
    mark size=1pt,
    line width=0.8pt,
    xlabel style={at={(axis description cs:0.5,-0.13)}, anchor=north},
    ylabel style={at={(axis description cs:-0.1,0.5)}, anchor=south},
]
\addplot+[mark=o, color=kl_green, thick] coordinates {
(0, 72.42) (100, 62.76) (200, 36.9) (300, 27.82) (400, 31.16) (500, 26.02) (600, 21.7) (700, 20.6) (800, 25.3) (900, 25.86) (1000, 19.9) (1100, 19.6) (1200, 21.62) (1300, 22.46) (1400, 22.9) (1500, 23.06) (1600, 17.92) (1700, 18.3)
};
\addplot+[mark=square*, color=kl_red, thick] coordinates {
(0, 72.42) (100, 74.7) (200, 20.62) (300, 6.5) (400, 1.26) (500, 1.1) (600, 0.24) (700, 0.6) (800, 0.32) (900, 0.4) (1000, 0.18) (1100, 0.1) (1200, 0.16) (1300, 0.1) (1400, 0.02) (1500, 0.1) (1600, 0.1) (1700, 0.08)  
};

% --- AP@50 Plot ---
\nextgroupplot[
    ylabel={AP@50},
    ymin=4, ymax=20,
    mark size=1pt,
    line width=0.8pt,
    xlabel style={at={(axis description cs:0.5,-0.13)}, anchor=north},
    ylabel style={at={(axis description cs:-0.1,0.5)}, anchor=south},
]
 
\addplot+[mark=o, color=kl_green, thick] coordinates {
(0, 8.10)  (100, 9.721) (200, 13.403) (300, 15.158) (400, 14.902) (500, 15.155) (600, 15.54) (700, 15.869) (800, 15.253) (900, 14.988) (1000, 15.667) (1100, 15.904) (1200, 15.733) (1300, 15.748) (1400, 15.37) (1500, 15.559) (1600, 16.1) (1700, 16.13)
};
\addplot+[mark=square*, color=kl_red, thick] coordinates {
(0, 8.10) (100, 6.5) (200, 14.641) (300, 16.847) (400, 17.246) (500, 17.345) (600, 17.811) (700, 17.525) (800, 17.537) (900, 17.669) (1000, 17.75) (1100, 17.634) (1200, 17.785) (1300, 17.787) (1400, 17.816) (1500, 17.872) (1600, 17.953) (1700, 18.075)
};

% --- Recall Plot ---
\nextgroupplot[
    ylabel={Recall (\%)},
    ymin=4, ymax=23,
    mark size=1pt,
    line width=0.8pt,
    xlabel style={at={(axis description cs:0.5,-0.13)}, anchor=north},
    ylabel style={at={(axis description cs:-0.1,0.5)}, anchor=south},
]
 
\addplot+[mark=o, color=kl_green, thick] coordinates {
(0, 5.47) (100, 7.56) (200, 11.26) (300, 13.1) (400, 12.91) (500, 13.81) (600, 14.79) (700, 15.61) (800, 14.29) (900, 14.11) (1000, 15.07) (1100, 15.56) (1200, 15.39) (1300, 15.03) (1400, 15.06) (1500, 15.12) (1600, 15.95) (1700, 16.08)
};
\addplot+[mark=square*, color=kl_red, thick] coordinates {
(0, 5.47)  (100, 5.77) (200, 14.27) (300, 17.86) (400, 18.72) (500, 19.42) (600, 18.43) (700, 19.22) (800, 19.59) (900, 19.5) (1000, 19.33) (1100, 20.04) (1200, 20.76) (1300, 20.75) (1400, 20.93) (1500, 21.04) (1600, 20.87) (1700, 21.05)  
};

%%%%%%%%%%%%% Sampling length %%%%%%%%%%%%%%%%
% --- Failure Rate Plot ---
\nextgroupplot[
    ylabel={Failure Rate (\%)},
    ymin=-5, ymax=100,
    mark size=1pt,
    line width=0.8pt,
    xlabel style={at={(axis description cs:0.5,-0.13)}, anchor=north},
    ylabel style={at={(axis description cs:-0.1,0.5)}, anchor=south},
]
 
\addplot+[mark=o, color=sample_blue, thick] coordinates {
(0, 72.42) (100, 97.8) (200, 94.44) (300, 74.96) (400, 55.78) (500, 25.82) (600, 38.62) (700, 28.44) (800, 28.86) (900, 30.4) (1000, 26.46) (1100, 21.26) (1200, 12) (1300, 14.56) (1400, 21.58) (1500, 8.3) (1600, 11.66) (1700, 13.62)
 };
\addplot+[mark=square*, color=sample_purple, thick] coordinates {
(0, 72.42) (100, 74.7) (200, 20.62) (300, 6.5) (400, 1.26) (500, 1.1) (600, 0.24) (700, 0.6) (800, 0.32) (900, 0.4) (1000, 0.18) (1100, 0.1) (1200, 0.16) (1300, 0.1) (1400, 0.02) (1500, 0.1) (1600, 0.1) (1700, 0.08)  
};

% --- AP@50 Plot ---
\nextgroupplot[
    ylabel={AP@50},
      ymin=0, ymax=20,
    mark size=1pt,
    line width=0.8pt,
    xlabel style={at={(axis description cs:0.5,-0.13)}, anchor=north},
    ylabel style={at={(axis description cs:-0.1,0.5)}, anchor=south},
]
\addplot+[mark=o, color=sample_blue, thick] coordinates {
(0, 8.10) (100, 1.104) (200, 1.674) (300, 5.463) (400, 7.952) (500, 12.198) (600, 10.539) (700, 11.821) (800, 11.924) (900, 11.164) (1000, 12.209) (1100, 12.777) (1200, 14.234) (1300, 13.699) (1400, 12.596) (1500, 14.792) (1600, 14.236) (1700, 14.171)
};
\addplot+[mark=square*, color=sample_purple, thick] coordinates {
(0, 8.10) (100, 6.5) (200, 14.641) (300, 16.847) (400, 17.246) (500, 17.345) (600, 17.811) (700, 17.525) (800, 17.537) (900, 17.669) (1000, 17.75) (1100, 17.634) (1200, 17.785) (1300, 17.787) (1400, 17.816) (1500, 17.872) (1600, 17.953) (1700, 18.075)
};

% --- Recall Plot ---
\nextgroupplot[
    ylabel={Recall (\%)},
     ymin=0, ymax=24,
    mark size=1pt,
    line width=0.8pt,
    xlabel style={at={(axis description cs:0.5,-0.13)}, anchor=north},
    ylabel style={at={(axis description cs:-0.1,0.5)}, anchor=south},
]
\addplot+[mark=o, color=sample_blue, thick] coordinates {
(0, 5.47) (100, 0.66) (200, 1.33) (300, 6.32) (400, 10.57) (500, 16.74) (600, 15.12) (700, 17.62) (800, 17.49) (900, 17.41) (1000, 18.27) (1100, 19.79) (1200, 22.06) (1300, 21.13) (1400, 19.6) (1500, 23) (1600, 21.96) (1700, 21.83)
};
\addplot+[mark=square*, color=sample_purple, thick] coordinates {
(0, 5.47)  (100, 5.77) (200, 14.27) (300, 17.86) (400, 18.72) (500, 19.42) (600, 18.43) (700, 19.22) (800, 19.59) (900, 19.5) (1000, 19.33) (1100, 20.04) (1200, 20.76) (1300, 20.75) (1400, 20.93) (1500, 21.04) (1600, 20.87) (1700, 21.05)  
};
%%%%%%%%%%%%% Group size %%%%%%%%%%%%%%%%
% --- Failure Rate Plot ---
\nextgroupplot[
    ylabel={Failure Rate (\%)},
    ymin=-10, ymax=100,
    mark size=1pt,
    line width=0.8pt,
    xlabel style={at={(axis description cs:0.5,-0.13)}, anchor=north, font=\scriptsize},
    ylabel style={at={(axis description cs:-0.1,0.5)}, anchor=south, font=\scriptsize},
]

\addplot+[mark=o, color=group_orange, thick] coordinates {
(0, 72.42) (100, 43.34) (200, 2.44) (300, 0.56) (400, 0.86) (500, 0.36) (600, 0.28) (700, 0.32) (800, 0.16) (900, 0.16) (1000, 0.18) (1100, 0.32) (1200, 0.14) (1300, 0.16) (1400, 0.08) (1500, 0.16) (1600, 0.18) (1700, 0.14)
};
%2b, g16
\addplot+[mark=square*, color=group_brown, thick] coordinates {
(0, 72.42) (100, 74.7) (200, 20.62) (300, 6.5) (400, 1.26) (500, 1.1) (600, 0.24) (700, 0.6) (800, 0.32) (900, 0.4) (1000, 0.18) (1100, 0.1) (1200, 0.16) (1300, 0.1) (1400, 0.02) (1500, 0.1) (1600, 0.1) (1700, 0.08)  
};

% --- AP@50 Plot ---
\nextgroupplot[
    ylabel={AP@50},
    ymin=4, ymax=20,
    mark size=1pt,
    line width=0.8pt,
    xlabel style={at={(axis description cs:0.5,-0.13)}, anchor=north, font=\scriptsize},
    ylabel style={at={(axis description cs:-0.1,0.5)}, anchor=south, font=\scriptsize},    
]
%g8
\addplot+[mark=o, color=group_orange, thick] coordinates {
(0, 8.10) (100, 11.77) (200, 17.08) (300, 17.426) (400, 17.653) (500, 17.578) (600, 17.734) (700, 17.868) (800, 18.039) (900, 18.251) (1000, 18.266) (1100, 18.275) (1200, 18.671) (1300, 18.295) (1400, 18.412) (1500, 18.378) (1600, 18.329) (1700, 18.389)
};
\addplot+[mark=square*, color=group_brown, thick] coordinates {
(0, 8.10) (100, 6.5) (200, 14.641) (300, 16.847) (400, 17.246) (500, 17.345) (600, 17.811) (700, 17.525) (800, 17.537) (900, 17.669) (1000, 17.75) (1100, 17.634) (1200, 17.785) (1300, 17.787) (1400, 17.816) (1500, 17.872) (1600, 17.953) (1700, 18.075)
};

% --- Recall Plot ---
\nextgroupplot[
    ylabel={Recall (\%)},
    ymin=4, ymax=23,
    mark size=1pt,
    line width=0.8pt,
    xlabel style={at={(axis description cs:0.5,-0.13)}, anchor=north, font=\scriptsize},
    ylabel style={at={(axis description cs:-0.1,0.5)}, anchor=south, font=\scriptsize},    
]
\addplot+[mark=o, color=group_orange, thick] coordinates {
(0, 5.47)  (100, 10.75) (200, 17.43) (300, 18.48) (400, 19.11) (500, 19.66) (600, 20.05) (700, 20.85) (800, 21.03) (900, 20.81) (1000, 20.98) (1100, 21.5) (1200, 21.65) (1300, 22.06) (1400, 21.87) (1500, 21.97) (1600, 21.96) (1700, 22.01)
};
\addplot+[mark=square*, color=group_brown, thick] coordinates {
(0, 5.47)  (100, 5.77) (200, 14.27) (300, 17.86) (400, 18.72) (500, 19.42) (600, 18.43) (700, 19.22) (800, 19.59) (900, 19.5) (1000, 19.33) (1100, 20.04) (1200, 20.76) (1300, 20.75) (1400, 20.93) (1500, 21.04) (1600, 20.87) (1700, 21.05) 
};

\end{groupplot}

% --- Unified Legend ---
\node at ($(group c1r1.north)!0.5!(group c3r1.north) + (0,2em)$) {%
\begin{tabular}{c c}
\tikz{\draw[kl_green, line width=0.8pt] (0,0) -- (0.7,0) node[pos=0.5, circle, draw=kl_green, fill=white, inner sep=1pt] {};} \, R1-SGG (2B,  \emph{w.} KL) 
%\hspace{1cm}
& 
\tikz{\draw[kl_red, line width=0.8pt] (0,0) -- (0.7,0) node[pos=0.5, rectangle, draw=kl_red, fill=white, inner sep=1pt] {};} \, R1-SGG (2B,  \emph{w/o} KL)
\end{tabular}
};
\node at ($(group c1r2.north)!0.5!(group c3r2.north) + (0,2em)$) {%
\begin{tabular}{c c}
\tikz{\draw[sample_blue, line width=0.8pt] (0,0) -- (0.7,0) node[pos=0.5, circle, draw=sample_blue, fill=white, inner sep=1pt] {};} \, R1-SGG (2B,  sampling length 2k) 
%\hspace{1cm}
& 
\tikz{\draw[sample_purple, line width=0.8pt] (0,0) -- (0.7,0) node[pos=0.5, rectangle, draw=sample_purple, fill=white, inner sep=1pt] {};} \, R1-SGG (2B,  sampling length 1k)
\end{tabular}
};
\node at ($(group c1r3.north)!0.5!(group c3r3.north) + (0,2em)$) {%
\begin{tabular}{c c}
\tikz{\draw[group_orange, line width=0.8pt] (0,0) -- (0.7,0) node[pos=0.5, circle, draw=group_orange, fill=white, inner sep=1pt] {};} \, R1-SGG (2B,  group size 16) 
%\hspace{1cm}
& 
\tikz{\draw[group_brown, line width=0.8pt] (0,0) -- (0.7,0) node[pos=0.5, rectangle, draw=group_brown, fill=white, inner sep=1pt] {};} \, R1-SGG (2B,  group size 8)
\end{tabular}
};

\end{tikzpicture}
}

\caption{
Performance comparison of R1-SGG (2B) across training steps on the VG150 validation set. 
Each row evaluates a different setting: (Top) KL divergence regularization ($\beta{=}0.04$ vs. $\beta{=}0$), (Middle) sampling length, and (Bottom) group size.
Metrics reported include Failure Rate (\%), AP@50, and Recall (\%).
}
\label{fig:kl_sample_group_vs_steps}
\end{figure*}

%%%%%%%%%%%%%%%%%%

%%%%%%%%% CHECK LIST
%%%%%%%%%%%%%%%%%%%%%%%%%%%%%%%%%%%%%%%%%%%%%%%%%%%%%%%%%%%%
%\input{checklist}

\end{document}